\documentclass{article}

\usepackage{microtype}
\usepackage{graphicx}
\usepackage{subfigure}
\usepackage{booktabs} 

\usepackage{hyperref}
\usepackage[accepted]{icml2019}

\usepackage{amsmath}
\usepackage{amssymb}
\usepackage{amsthm}
\usepackage{bm}
\usepackage{times}
\usepackage{graphicx} 
\usepackage{subfigure} 
\usepackage{multirow}
\usepackage{color}
\usepackage{threeparttable}
\usepackage{algorithm}
\usepackage{algorithmic}
\usepackage{multirow}

\newtheorem{theorem}{Theorem}[section]

\newtheorem{proposition}[theorem]{Proposition}

\newtheorem{definition}[theorem]{Definition}

\icmltitlerunning{Gromov-Wasserstein Learning for Graph Matching and Node Embedding}

\begin{document}

\twocolumn[
\icmltitle{Gromov-Wasserstein Learning for Graph Matching and Node Embedding}



\icmlsetsymbol{equal}{*}

\begin{icmlauthorlist}
\icmlauthor{Hongteng Xu}{in,du}
\icmlauthor{Dixin Luo}{du}
\icmlauthor{Hongyuan Zha}{gt}
\icmlauthor{Lawrence Carin}{du}
\end{icmlauthorlist}

\icmlaffiliation{in}{Infinia ML, Inc., Durham, NC, USA}
\icmlaffiliation{du}{Department of ECE, Duke University, Durham, NC, USA}
\icmlaffiliation{gt}{College of Computing, Georgia Institute of Technology, Atlanta, GA, USA}
\icmlcorrespondingauthor{Hongteng Xu}{hongtengxu313@gmail.com}

\icmlkeywords{Gromov-Wasserstein discrepancy, optimal transport, node embedding, graph matching}

\vskip 0.3in
]



\printAffiliationsAndNotice{}  

\begin{abstract}
A novel Gromov-Wasserstein learning framework is proposed to jointly match (align) graphs and learn embedding vectors for the associated graph nodes. 
Using Gromov-Wasserstein discrepancy, we measure the dissimilarity between two graphs and find their correspondence, according to the learned optimal transport. 
The node embeddings associated with the two graphs are learned under the guidance of the optimal transport, the distance of which not only reflects the topological structure of each graph but also yields the correspondence across the graphs. 
These two learning steps are mutually-beneficial, and are unified here by minimizing the Gromov-Wasserstein discrepancy with structural regularizers. 
This framework leads to an optimization problem that is solved by a proximal point method.
We apply the proposed method to matching problems in real-world networks, and demonstrate its superior performance compared to alternative approaches.
\end{abstract}

\section{Introduction}
Real-world entities and their interactions are often represented as graphs.
Given two or more graphs created in different domains, graph matching aims to find a correspondence across different graphs. 
This task is important for many applications, $e.g.$, matching the protein networks from different species~\cite{sharan2006modeling,singh2008global}, linking accounts in different social networks~\cite{zhang2015multiple}, and feature matching in computer vision~\cite{cordella2004sub}.
However, because it is NP-hard, graph matching is challenging and often solved heuristically. 
Further complicating matters, the observed graphs may be noisy ($e.g.$, containing unreliable edges), which leads to unsatisfying matching results using traditional methods. 

\begin{figure}[t]
    \centering
    \includegraphics[width=0.85\linewidth]{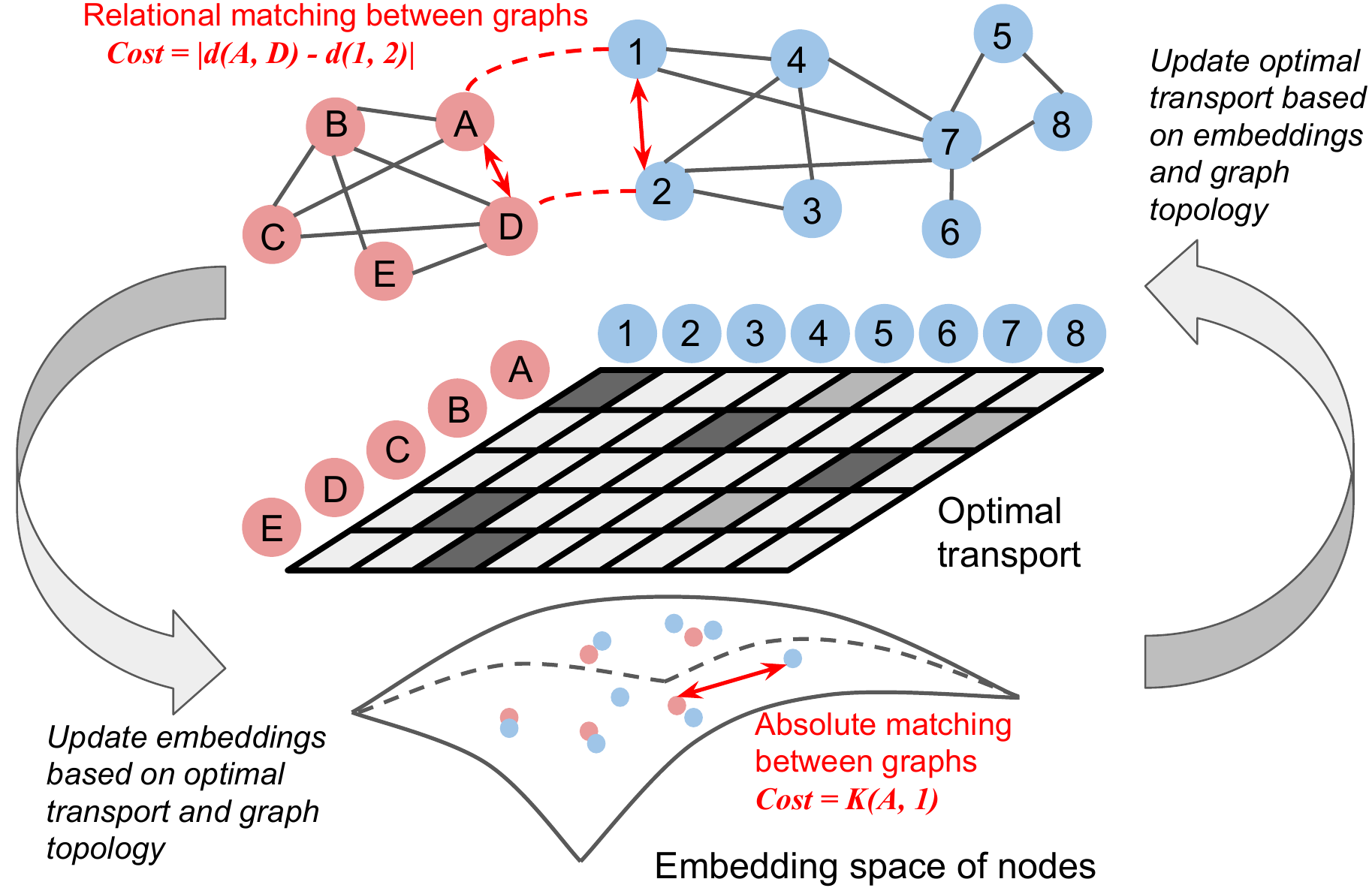}
    \vspace{-11pt}
    \caption{An illustration of the proposed method.}\vspace{-15pt}
    \label{fig:scheme}
\end{figure}

A problem related to graph matching is the learning of node embeddings, which aims to learn a latent vector for each graph node; the collection of embeddings approximates the topology of the graph, with similar/related nodes nearby in embedding space. 
Learning suitable node embeddings is beneficial for graph matching, as one may seek to align two or more graphs according to the metric structure associated with their node embeddings. 
Although graph matching and node embedding are highly related tasks, in practice they are often treated and solved independently. 
Existing node embedding methods~\cite{perozzi2014deepwalk,tang2015line,grover2016node2vec} are designed for a single graph, and applying such methods separately to multiple graphs doesn't share information across the graphs, and, hence, is less helpful for graph matching. 
Most graph matching methods rely purely on topological information ($i.e.$, adjacency matrices of graphs) and ignore the potential functionality of node embeddings~\cite{kuchaiev2010topological,neyshabur2013netal,nassar2018low}. 
Although some methods consider first deriving embeddings for each graph and then learning a transformation between the embeddings, their results are often unsatisfying because their embeddings are predefined and the transformations are limited to orthogonal projections~\cite{grave2018unsupervised} or rigid/non-rigid deformations~\cite{myronenko2010point}. 

This paper considers the joint goal of graph matching and learning node embeddings, seeking to achieve improvements in both tasks. 
As illustrated in Figure~\ref{fig:scheme}, to achieve this goal we propose a novel Gromov-Wasserstein learning framework. 
The dissimilarity between two graphs is measured by the Gromov-Wasserstein discrepancy (GW discrepancy)~\cite{peyre2016gromov}, which compares the distance matrices of different graphs in a relational manner, and learns an optimal transport between the nodes of different graphs. 
The learned optimal transport indicates the correspondence between the graphs.
The embeddings of the nodes from different graphs are learned jointly: the distance between the embeddings within the same graph should approach the distance matrix derived from data, and the distance between the embeddings across different graphs should reflect the correspondence indicated by the learned optimal transport. 
As a result, the objectives of graph matching and node embedding are unified as minimizing the Gromov-Wasserstein discrepancy between two graphs, with structural regularizers. 
This framework leads to an optimization problem that is solved via an iterative process. 
In each iteration, the embeddings are used to estimate distance matrices when learning the optimal transport, and the learned optimal transport regularizes the learning of embeddings in the next iteration.

There are two important benefits to tackling graph matching and node embedding jointly. 
First, the observed graphs often contain spurious edges or miss some useful edges, leading to noisy adjacency matrices and unreliable graph matching results. 
Treating the distance between learned node embeddings as complementary information of observed edges, we can approximate the topology of graph more robustly, and accordingly, match noisy graphs.
Second, our method regularizes the GW discrepancy and learns embeddings of different graphs on the same manifold,
instead of learning an explicit transformation between the embeddings with predefined constraints. 
Therefore, the proposed method is more flexible and has lower risk of model misspecification ($i.e.$, imposing incorrect constraints on the transformation); the distance between the embeddings of different graphs can be calculated directly without any additional transformation. 
We test our method on real-world matching problems and analyze its performance in depth. 
Experiments show that our method obtains encouraging matching results, with comparisons made to alternative approaches.

\section{Gromov-Wasserstein Learning Framework}\label{sec:method}
Assume we have two sets of entities (nodes), denoted as source set $\mathcal{V}_s$ and target set $\mathcal{V}_t$. 
Without loss of generality, we assume that $|\mathcal{V}_s|\leq |\mathcal{V}_t|$.
For each set, we observe a set of interactions between its entities, $i.e.$, $\mathcal{E}_k=\{(v_i, v_j, w_{ij}) | v_i, v_j \in \mathcal{V}_k\}$, where $k=s$ or $t$, and $w_{ij}$ counts the appearances of the interaction $(v_i, v_j)$. 
Accordingly, the data of these entities can be represented as two graphs, denoted as $G(\mathcal{V}_s, \mathcal{E}_s)$ and $G(\mathcal{V}_t, \mathcal{E}_t)$, and we focus on the following two tasks: 
$i$) Find a correspondence between the graphs. 
$ii$) Obtain node embeddings of the two graphs, $i.e.$, $\bm{X}_s=[\bm{x}_i^{s}]\in\mathbb{R}^{D\times |\mathcal{V}_s|}$ and $\bm{X}_t=[\bm{x}_i^{t}]\in\mathbb{R}^{D\times |\mathcal{V}_t|}$. 
These two tasks are unified in a framework based on Gromov-Wasserstein discrepancy.

\subsection{Gromov-Wasserstein discrepancy}
Gromov-Wasserstein discrepancy was proposed in~\cite{peyre2016gromov}, which is a natural extension of Gromov-Wasserstein distance~\cite{memoli2011gromov}. 
Specifically, the definition of Gromov-Wasserstein distance is as follows:
\begin{definition}
Let $(X, d_X, \mu_X)$ and $(Y, d_Y, \mu_Y)$ be two metric measure spaces, where $(X, d_X)$ is a compact metric space and $\mu_X$ is a Borel probability measure on $X$ (with $(Y, d_Y, \mu_Y)$ defined in the same way). 
The Gromov-Wasserstein distance $d_{GW}(\mu_X, \mu_Y)$ is defined as
\begin{eqnarray*}
\inf_{\pi\in \Pi(\mu_X, \mu_Y)}\iint\limits_{X\times Y, X\times Y}L(x, y, x', y')\,d\pi(x, y)\,d\pi(x',y'),
\end{eqnarray*}
where $L(x, y, x', y')=|d_X(x, x')-d_Y(y,y')|$ is the loss function and $\Pi(\mu_X,\mu_Y)$ is the set of all probability measures on $X\times Y$ with $\mu_X$ and $\mu_Y$ as marginals.
\end{definition}
This defines an optimal transport-like distance~\cite{villani2008optimal} for metric spaces: it calculates distances between pairs of samples within each domain and measures how these distances compare to those in the other domain. 
It does not require one to directly compare the samples across different spaces and the target spaces can have different dimensions.
When $d_X$ and $d_Y$ are replaced with dissimilarity measurements rather than strict distance metrics, and the loss function $L$ is defined more flexibly, $e.g.$, mean-square-error (MSE) or KL-divergence, we relax the Gromov-Wasserstein distance to the proposed Gromov-Wasserstein {\em discrepancy}.
These relaxations make the proposed Gromov-Wasserstein learning framework suitable for a wide range of machine learning tasks, including graph matching.

In graph matching, a metric-measure space corresponds to the pair $(\bm{C},\bm{\mu})\in \mathbb{R}^{|\mathcal{V}|\times|\mathcal{V}|}\times \Sigma^{|\mathcal{V}|}$ of a graph $G(\mathcal{V},\mathcal{E})$, where $\bm{C}=[c_{ij}]\in\mathbb{R}^{|\mathcal{V}|\times |\mathcal{V}|}$ represents a distance/dissimilarity matrix derived according to the interaction set $\mathcal{E}$, $i.e.$, each $c_{ij}$ is a function of $w_{ij}$. 
The empirical distribution of nodes, $i.e.$, $\bm{\mu}=[\mu_i]\in\Sigma^{|\mathcal{V}|}$, is calculated based on the normalized degree of graph.
It reflects the probability of each node appearing in observed interactions.
Given two graphs $G(\mathcal{V}_s,\mathcal{E}_s)$ and $G(\mathcal{V}_t,\mathcal{E}_t)$, the Gromov-Wasserstein discrepancy between $(\bm{C}_s,\bm{\mu}_s)$ and $(\bm{C}_t,\bm{\mu}_t)$ is defined as
\begin{eqnarray}\label{eq:dgw}
\begin{aligned}
&d_{GW}(\bm{\mu}_s,\bm{\mu_t})\\
&:=\sideset{}{_{\bm{T}\in\Pi(\bm{\mu}_s,\bm{\mu_t})}}\min\sideset{}{_{i,j,i',j'}}\sum L(c_{ij}^s,c_{i'j'}^t)T_{ii'}T_{jj'}\\
&=\sideset{}{_{\bm{T}\in\Pi(\bm{\mu}_s,\bm{\mu_t})}}\min\langle \bm{L}(\bm{C}_s,\bm{C}_t, \bm{T}), \bm{T}\rangle.
\end{aligned}
\end{eqnarray}
Here, $\Pi(\bm{\mu}_s,~\bm{\mu}_t)=\{\bm{T}\in\mathbb{R}^{|\mathcal{V}_s|\times |\mathcal{V}_t|}~|~\bm{T}\bm{1}_{|\mathcal{V}_t|}=\bm{\mu}_s,~\bm{T}^{\top}\bm{1}_{|\mathcal{V}_s|}=\bm{\mu}_t\}$. 
$L(\cdot,\cdot)$ is an element-wise loss function, with typical choices the square loss $L(a, b)=(a-b)^2$ and the KL-divergence $L(a, b)=a\log\frac{a}{b}-a+b$. 
Accordingly, $\bm{L}(\bm{C}_s,\bm{C}_t, \bm{T})=[L_{jj'}]\in\mathbb{R}^{|\mathcal{V}_s|\times |\mathcal{V}_t|}$ and each $L_{jj'}=\sum_{i,i'}L(c_{ij}^s,c_{i'j'}^t)T_{ii'}$, and $\langle\cdot,\cdot\rangle$ represents the inner product of matrices; $\bm{T}$ is the optimal transport between the nodes of two graphs, and its element $T_{ij}$ represents the probability that $v_{i}\in\mathcal{V}_s$ matches $v_j\in\mathcal{V}_t$. 
By choosing the largest $T_{ij}$ for each $i$, we find the correspondence that minimizes the GW discrepancy between the two graphs. 

However, such a graph matching strategy raises several issues. 
First, for each graph, its observed interaction set can be noisy, which leads to an unreliable distance matrix. 
Minimizing the GW discrepancy based on such distance matrices has a negative influence on matching results. 
Second, the Gromov-Wasserstein discrepancy compares different graphs relationally based on their edges ($i.e.$, the distance between a pair of nodes within each graph), while most existing graph matching methods consider the information of nodes {\em and} edges jointly~\cite{neyshabur2013netal,vijayan2015magna++,sun2015simultaneous}. 
Therefore, to make a successful graph matching method, we further consider the learning of node embeddings and derive the proposed framework. 

\subsection{Proposed model}
We propose to not only learn the optimal transport indicating the correspondence between graphs but also simultaneously learn the node embeddings for each graph, which leads to a regularized Gromov-Wasserstein discrepancy. 
The corresponding optimization problem is
\begin{eqnarray}\label{eq:rgwl}
\begin{aligned}
&\min_{\bm{X}_s,\bm{X}_t}~\min_{\bm{T}\in \Pi(\bm{\mu}_s,\bm{\mu}_t)}\underbrace{\langle \bm{L}(\bm{C}_s(\bm{X}_s), \bm{C}_t(\bm{X}_t),\bm{T}),~\bm{T} \rangle}_{\text{Gromov-Wasserstein discrepancy}} \\
&\quad\quad\quad\quad + \underbrace{\alpha\langle \bm{K}(\bm{X}_s, \bm{X}_t),~\bm{T} \rangle}_{\text{Wasserstein discrepancy}}
+ \underbrace{\beta R(\bm{X}_s,\bm{X}_t)}_{\text{prior information}}.
\end{aligned}
\end{eqnarray}
The first term in (\ref{eq:rgwl}) corresponds to the GW discrepancy defined in (\ref{eq:dgw}), which measures the \emph{relational dissimilarity} between the two graphs. 
The difference here is that the proposed distance matrices consider both the information of observed data and that of embeddings:
\begin{eqnarray}\label{eq:cost}
\begin{aligned}
\bm{C}_k(\bm{X}_k) = (1-\alpha)\bm{C}_k + \alpha\bm{K}(\bm{X}_k, \bm{X}_k),~\mbox{for}~k=s,t.
\end{aligned}
\end{eqnarray}
Here $\bm{K}(\bm{X}_k,\bm{X}_k)=[\kappa(\bm{x}_i^k,\bm{x}_{j}^k)]\in\mathbb{R}^{|\mathcal{V}_k\times |\mathcal{V}_k|}$ is a distance matrix, with element $\kappa(\bm{x}_i^k,\bm{x}_j^k)$ that is a function measuring the distance between the node embeddings within the same graph;
$\alpha\in [0, 1]$ is a hyperparameter controlling the contribution of embedding-based distance to $\bm{C}_k(\bm{X}_k)$. 

The second term in (\ref{eq:rgwl}) represents the Wasserstein discrepancy between the nodes of the two graphs. 
Similar to the first term, the distance matrix is also derived based on the node embeddings, $i.e.$, $\bm{K}(\bm{X}_s,\bm{X}_t)=[\kappa(\bm{x}_i^s,\bm{x}_j^t)]\in\mathbb{R}^{|\mathcal{V}_s|\times |\mathcal{V}_t|}$, and its contribution is controlled by the same hyperparameter $\alpha$.
This term measures the \emph{absolute dissimilarity} between the two graphs, which connects the target optimal transport with node embeddings. 
By adding this term, the optimal transport minimizes both the Gromov-Wasserstein discrepancy based directly on observed data and the Wasserstein discrepancy based on the embeddings (which are indirectly also a function of the data). 
Furthermore, the embeddings of different graphs can be learned jointly under the guidance of the optimal transport --- the distance between the embeddings of different graphs should be consistent with the relationship indicated by the optimal transport. 

Because the target optimal transport is often sparse, purely considering its guidance leads to overfitting or trivial solutions when learning embeddings. 
To mitigate this problem, the third term in (\ref{eq:rgwl}) represents a regularization of the embeddings, based on the prior information provided by $\bm{C}_s$ and $\bm{C}_t$. 
We require the embedding-based distance matrices to be close to the observed ones, and $R(\bm{X}_s,\bm{X}_t)$ is 
\begin{eqnarray}\label{eq:reg}
\begin{aligned}
\sideset{}{_{k=s,t}}\sum L(\bm{K}(\bm{X}_k,\bm{X}_k), \bm{C}_k)+\underbrace{L(\bm{K}(\bm{X}_s,\bm{X}_t), \bm{C}_{st})}_{\text{optional}},
\end{aligned}
\end{eqnarray}
where the definition of loss function $L(\cdot, \cdot)$ is the same as that used in (\ref{eq:dgw}). 
Note that if we observe partial correspondences between different graphs, $i.e.$, $\mathcal{E}_{st}=\{(v_i, v_j, w_{ij})|v_i\in \mathcal{V}_s, v_j\in\mathcal{V}_t\}$, we can calculate a distance matrix for the nodes of different graphs, denoted as $\bm{C}_{st}\in \mathbb{R}^{|\mathcal{V}_s|\times |\mathcal{V}_t|}$, and require the distance between the embeddings to match with $\bm{C}_{st}$, as shown in the optional term of (\ref{eq:reg}). 
This term is available only when $\mathcal{E}_{st}$ is given. 

The proposed method unifies (optimal transport-based) graph matching and node embedding in the same framework, and makes them beneficial to each other. 
For the original GW discrepancy term, introducing the embedding-based distance matrices can suppress the noise in the data-driven distance matrices, improving robustness. 
Additionally, based on node embeddings, we can calculate the Wasserstein discrepancy between graphs, which further regularizes the target optimal transport directly. 
When learning node embeddings, the Wasserstein discrepancy term works as the regularizer of node embeddings --- the values of the learned optimal transport indicate which pairs of nodes should be close to each other.

\section{Learning Algorithm}
\subsection{Learning optimal transport}
Although (\ref{eq:rgwl}) is a complicated nonconvex optimization problem, we can solve it effectively by alternatively learning the optimal transport and the embeddings. 
In particular, the proposed method applies nested iterative optimization. 
In the $m$-th outer iteration, given current embeddings $\bm{X}_s^{(m)}$ and $\bm{X}_t^{(m)}$, we solve the following sub-problem:
\begin{eqnarray}\label{eq:ot}
\begin{aligned}
\sideset{}{_{\bm{T}\in \Pi(\bm{\mu}_s,\bm{\mu}_t)}}\min&\langle \bm{L}(\bm{C}_s(\bm{X}_s^{(m)}), \bm{C}_t(\bm{X}_t^{(m)}),\bm{T}),~\bm{T} \rangle\\
&+\alpha \langle\bm{K}(\bm{X}_s^{(m)}, \bm{X}_t^{(m)}),~\bm{T} \rangle.
\end{aligned}
\end{eqnarray}
This sub-problem is still nonconvex because of the quadratic term $\bm{L}(\bm{C}_s(\bm{X}_s^{(m)}), \bm{C}_t(\bm{X}_t^{(m)}),\bm{T}),~\bm{T} \rangle$.
We solve it iteratively with the help of a proximal point method. 
Inspired by the method in~\cite{xie2018fast}, in the $n$-th inner iteration we update the target optimal transport via 
\begin{eqnarray}\label{eq:ipot}
\begin{aligned}
&\sideset{}{_{\bm{T}\in \Pi(\bm{\mu}_s,\bm{\mu}_t)}}\min\langle \bm{L}(\bm{C}_s(\bm{X}_s^{(m)}), \bm{C}_t(\bm{X}_t^{(m)}),\bm{T}),\bm{T} \rangle\\
&\quad\quad+\alpha\langle \bm{K}(\bm{X}_s^{(m)}, \bm{X}_t^{(m)}),~\bm{T} \rangle+\gamma \mbox{KL}(\bm{T}\lVert \bm{T}^{(n)}).
\end{aligned}
\end{eqnarray}
Here, a proximal term based on Kullback-Leibler (KL) divergence, $\mbox{KL}(\bm{T}\lVert \bm{T}^{(n)})=\sum_{ij}T_{ij}\log \frac{T_{ij}}{T_{ij}^{(n)}} - T_{ij} +T_{ij}^{(n)}$, is added as a regularizer.

We use projected gradient descent to solve (\ref{eq:ipot}), in which both the gradient and the projection are based on the KL metric. 
When the learning rate is set as $\frac{1}{\gamma}$, the projected gradient descent is equivalent to solving the following optimal transport problem with an entropy regularizer~\cite{benamou2015iterative,peyre2016gromov}:
\begin{eqnarray}\label{eq:sinkhorn}
\begin{aligned}
\sideset{}{_{\bm{T}\in \Pi(\bm{\mu}_s,\bm{\mu}_t)}}\min\langle \bm{C}^{(m,n)}-\gamma\log\bm{T}^{(n)},\bm{T} \rangle+\gamma H(\bm{T}),
\end{aligned}
\end{eqnarray}
where $\bm{C}^{(m,n)}=\bm{L}(\bm{C}_s, \bm{C}_t,\bm{T}^{(n)})+\alpha \bm{K}(\bm{X}_s^{(m)}, \bm{X}_t^{(m)})+\gamma$, and $H(\bm{T})=\sum_{i,j}T_{ij}\log T_{ij}$. 
This problem can be solved via the Sinkhorn-Knopp algorithm~\cite{sinkhorn1967concerning,cuturi2013sinkhorn} with linear convergence.

In summary, we decompose (\ref{eq:ot}) into a series of updating steps.  Each updating step (\ref{eq:ipot}) can be solved via projected gradient descent, which is a solution to a regularized optimal transport problem (\ref{eq:sinkhorn}). 
Essentially, the proposed method can be viewed as a special case of successive upper-bound minimization (SUM)~\cite{razaviyayn2013unified}, whose global convergence is guaranteed:
\begin{proposition}\label{prop}
Every limit point generated by our proximal point method, $i.e.$, $\lim_{n\rightarrow\infty}\bm{T}^{(n)}$, is a stationary point of the problem (\ref{eq:ot}).
\end{proposition}

Note that besides our proximal point method, another method for solving (\ref{eq:ot}) involves replacing the KL-divergence $\mbox{KL}(\bm{T}\lVert \bm{T}^{(n)})$ in (\ref{eq:ipot}) with an entropy regularizer $H(\bm{T})$ and minimizing an entropic GW discrepancy via iterative Sinkhorn projection~\cite{peyre2016gromov}. 
However, its performance ($i.e.$, its convergence and numerical stability) is more sensitive to the choice of the hyperparameter $\gamma$.
The details of our proximal point method, the proof of Proposition~\ref{prop}, and its comparison with the Sinkhorn method~\cite{peyre2016gromov} are shown in the Supplementary Material.

Parameter $\alpha$ controls the influence of node embeddings on the GW discrepancy and the Wasserstein discrepancy. 
When training the proposed model from scratch, the embeddings $\bm{X}_s$ and $\bm{X}_t$ are initialized randomly and thus are unreliable in the beginning. 
Therefore, we initialize $\alpha$ with a small value and increase it with respect to the number of outer iterations. 
We apply a simple linear strategy to adjust $\alpha$: with the maximum number of outer iterations set as $M$, in the $m$-th iteration, we set $\alpha_m = \frac{m}{M}$. 

\subsection{Updating embeddings}
Given the optimal transport, $\widehat{\bm{T}}^{(m)}$, we update the embeddings by solving the following optimization problem:
\begin{eqnarray}\label{eq:emb}
\begin{aligned}
\sideset{}{_{\bm{X}_s,\bm{X}_t}}\min &\alpha_m\langle \bm{K}(\bm{X}_s, \bm{X}_t),~\widehat{\bm{T}}^{(m)} \rangle
+ \beta R(\bm{X}_s,\bm{X}_t).
\end{aligned}
\end{eqnarray}
This problem can be solved effectively by (stochastic) gradient descent. 
In summary, the proposed learning algorithm is shown in Algorithm~\ref{alg2}.
\begin{algorithm}[t]
	\caption{Gromov-Wasserstein Learning (GWL)}
	\label{alg2}
	\begin{algorithmic}[1]
		\STATE \textbf{Input:} $\{\bm{C}_s,\bm{C}_t\}$, $\{\bm{\mu}_s,\bm{\mu}_t\}$, $\beta$, $\gamma$, 
		the dimension $D$, the number of outer/inner iterations $\{M,N\}$.
		\STATE \textbf{Output:} $\bm{X}_s$, $\bm{X}_t$ and $\widehat{\bm{T}}$.
		\STATE Initialize $\bm{X}_s^{(0)}$, $\bm{X}_t^{(0)}$ randomly, $\widehat{\bm{T}}^{(0)}=\bm{\mu}_s\bm{\mu}_t^{\top}$.
		\STATE{\textbf{For} $m=0:M-1$}
		\STATE \quad Set $\alpha_m = \frac{m}{M}$.
		\STATE{\quad\textbf{For} $n=0:N-1$}
		\STATE \quad\quad Update optimal transport $\widehat{T}^{(m+1)}$ via solving~(\ref{eq:ipot}).
		\STATE \quad Obtain $\bm{X}_s^{(m+1)}$, $\bm{X}_t^{(m+1)}$ via solving~(\ref{eq:emb}).
		\STATE $\bm{X}_s=\bm{X}_s^{(M)}$, $\bm{X}_t=\bm{X}_t^{(M)}$ and $\widehat{\bm{T}}=\widehat{\bm{T}}^{(M)}$.
	    \STATE $\backslash\backslash$ \texttt{Graph matching:}
	    \STATE Initialize correspondence set $\mathcal{P}=\emptyset$
	    \STATE{\textbf{For} $v_i\in\mathcal{V}_s$}
	    \STATE \quad $j=\arg\max_{j}\widehat{T}_{ij}$. $\mathcal{P}=\mathcal{P}\cup \{(v_i\in\mathcal{V}_s, v_j\in\mathcal{V}_t)\}$.
	\end{algorithmic}
\end{algorithm}

\subsection{Implementation details and analysis}
\textbf{Distance matrix}
The distance matrix plays an important role in our Gromov-Wasserstein learning framework. 
For a graph, the data-driven distance matrix should reflect its structure. 
Based on the fact that the counts of interactions in many real-world graphs is characterized by Zipf's law~\cite{powers1998applications}, we treat the counts as the weights of edges and define the element of the data-driven distance matrix as
\begin{eqnarray}\label{eq:zipf}
\begin{aligned}
c_{ij}^{k} = 
\begin{cases}
\frac{1}{w_{ij}+1}, & (v_i, v_j)\in\mathcal{E}_k,\\
1, & (v_i, v_j)\notin\mathcal{E}_k,
\end{cases}
~~\mbox{for}~k=s,t.
\end{aligned}
\end{eqnarray}
This definition assigns a short distance to pairs of nodes with many interactions. 
Additionally, we hope that the embedding-based distance matrix can fit the data-driven distance matrix easily. 
In the following experiments, we test two kinds of embedding-based distance: 
1) Cosine-based distance:
$\kappa(\bm{x}_i, \bm{x}_j) = 1 - \exp(-\sigma(1-\frac{\bm{x}_i^{\top}\bm{x}_j}{\|\bm{x}_i\|_2\|\bm{x}_j\|_2}))$. 
2) Radial basis function (RBF)-based distance:
$\kappa(\bm{x}_i, \bm{x}_j) = 1 - \exp(-\frac{\|\bm{x}_i-\bm{x}_j\|_2^2}{\sigma^2})$. 
When applying the cosine-based distance, we choose $\sigma=10$ such that the maximum $\kappa(\bm{x}_i, \bm{x}_j)$ approaches to $1$.
When applying the RBF-based distance, we choose $\sigma=D$. 
The following experiments show that these two distances work well in various matching tasks.

\textbf{Complexity and Scalability}
When learning optimal transport, one of the most time-consuming steps is computing the loss matrix $\bm{L}(\bm{C}_s, \bm{C}_t, \bm{T})$, which involves a tensor-matrix multiplication. 
Fortunately, as shown in~\cite{peyre2016gromov} when the loss function $L(a, b)$ can be written as $L(a, b)=f_1(a)+f_2(b)-h_1(a)h_2(b)$ for functions $(f_1,f_2, h_1, h_2)$, which is satisfied by our MSE/KL loss, the loss matrix can be calculated as $\bm{L}(\bm{C}_s, \bm{C}_t, \bm{T})=f_1(\bm{C}_s)\bm{\mu}_s\bm{1}_{|\mathcal{V}_t|}^{\top}+\bm{1}_{|\mathcal{V}_s|}\bm{\mu}_t^{\top}f_2(\bm{C}_t)^{\top}-h_1(\bm{C}_s)\bm{T}h_2(\bm{C}_t)^{\top}$.
Because $\bm{T}$ tends to be sparse quickly during the learning process, the computational complexity of $L(a, b)=f_1(a)+f_2(b)-h_1(a)h_2(b)$ is $\mathcal{O}(V^3)$, where $V=\max\{|\mathcal{V}_s|,|\mathcal{V}_t|\}$. 
For $D$-dimensional node embeddings, the complexity of the embedding-based distance matrix $\bm{K}(\bm{X}_s,\bm{X}_t)$ is $\mathcal{O}(V^2D)$.
Additionally, we can apply the inexact proximal point method~\cite{xie2018fast,chen2018adversarial}, running one-step Sinkhorn-Knopp projection in each inner iteration. 
Therefore, the complexity of learning optimal transport is $\mathcal{O}(V^2D + NV^3)$. 
When learning node embeddings, we can apply stochastic gradient descent to solve (\ref{eq:emb}). 
In our experiments, we select the size of the node batch as $B\ll V$ and the objective function of (\ref{eq:emb}) converges quickly after a few epochs.  Therefore, the computational complexity of the embedding-based distance sub-matrix is just $\mathcal{O}(B^2D)$, which may be ignored compared to that of learning optimal transport. 
In summary, the overall complexity of our method is $\mathcal{O}(M(V^2D + NV^3))$, and both the learning of optimal transport and that of node embeddings can be done in parallel on GPUs. 

Note that the proposed method has lower complexity than many existing graph matching methods. 
For example, the GRAAL and its variants~\cite{malod2015graal} have $\mathcal{O}(V^5)$ complexity, which is much slower than the proposed  method. 
Additionally, the complexity of our method is independent of the number of edges (denoted as $E=\max\{|\mathcal{E}_s|,|\mathcal{E}_t|\}$). 
Compared to other well-known alternatives, $e.g.$, NETAL~\cite{neyshabur2013netal} with $\mathcal{O}(V\log V + E^2 + EV\log V)$, our method has at least comparable complexity for dense graphs ($E\gg V$).

\section{Related Work}
\textbf{Gromov-Wasserstein learning} 
Gromov-Wasserstein discrepancy extends optimal transport~\cite{villani2008optimal} to the case when the target domains are not registered well.
It can also be viewed as a relaxation of Gromov-Hausdorff distance~\cite{memoli2008gromov,bronstein2010gromov} when pairwise distance between entities is defined. 
The GW discrepancy is suitable for solving matching problems like shape and object matching~\cite{memoli2009spectral,memoli2011gromov}. 
Besides graphics and computer vision, recently its potential for other applications has been investigated, $e.g.$, matching vocabulary sets between different languages~\cite{alvarez2018gromov} and matching weighted directed networks~\cite{chowdhury2018gromov}. 
The work in~\cite{peyre2016gromov} considers the Gromov-Wasserstein barycenter and proposes a fast Sinkhorn projection-based algorithm to compute GW discrepancy~\cite{cuturi2013sinkhorn}. 
Similar to our method, the work in~\cite{vayer2018fused} proposes a fused Gromov-Wasserstein distance, combining GW discrepancy with Wasserstein discrepancy.
However, it does not consider the learning of embeddings and requires the distance between the entities in different domains to be known, which is inapplicable to matching problems. 
In~\cite{bunne2018}, an adversarial learning method is proposed to learn a pair of generative models for incomparable spaces, which uses GW discrepancy as the objective function. 
This method imposes an orthogonal assumption on the transformation between the sample and its embedding; 
it is designed for fuzzy matching between distributions, rather than the graph matching task that requires point-to-point correspondence.

\textbf{Graph matching}
Graph matching has been studied extensively, with a wide range of applications. 
Focusing on protein-protein interaction (PPI) networks, many methods have been proposed, including methods based on local neighborhood information like GRAAL~\cite{kuchaiev2010topological}, and its variants MI-GRAAL~\cite{kuchaiev2011integrative} and L-GRAAL~\cite{malod2015graal}; as well as methods based on global structural information, like IsoRank~\cite{singh2008global}, MAGNA++~\cite{vijayan2015magna++}, NETAL~\cite{neyshabur2013netal}, HubAlign~\cite{hashemifar2014hubalign} and WAVE~\cite{sun2015simultaneous}. 
Among these methods, MAGNA++ and WAVE consider both edge and node information.
Besides bioinformatics, network alignment techniques are also applied to computer vision~\cite{jun2017sequential,yu2018generalizing}, document analysis~\cite{bayati2009algorithms} and social network analysis~\cite{zhang2015multiple}. 
For small graphs, $e.g.$, the graph of feature points in computer vision, graph matching is often solved as a quadratic assignment problem~\cite{yan2015matrix}. 
For large graphs, $e.g.$, social networks and PPI networks, existing methods either depend on a heuristic searching strategy or leverage domain knowledge for specific cases. 
{None of these methods consider graph matching and node embedding jointly from the viewpoint of Gromov-Wasserstein discrepancy.}

\textbf{Node embedding} Node embedding techniques have been widely used to represent and analyze graph/network structures.
The representative methods include LINE~\cite{tang2015line}, Deepwalk~\cite{perozzi2014deepwalk}, and node2vec~\cite{grover2016node2vec}. 
Most of these embedding methods first generate sequential observations of nodes through a random-walk procedure, and then learn the embeddings by maximizing the coherency between each observation and its context~\cite{mikolov2013efficient}. 
The distance between the learned embeddings can reflect the topological structure of the graph. 
More recently, many new embedding methods have been proposed, $e.g.$, the anonymous walk embedding in~\cite{ivanov2018anonymous} and the mixed membership word embedding~\cite{foulds2018mixed}, which help to improve the representations of complicated graphs and their nodes. 
However, none of these methods consider jointly learning embeddings for multiple graphs.

\section{Experiments}
We apply the Gromov-Wasserstein learning (GWL) method to both synthetic and real-world matching tasks, and compare it with state-of-the-art methods.
In our experiments, we set hyperparameters as follows: the number of outer iterations is $M=30$, the number of inner iteration is $N=200$, $\gamma=0.01$ and $L(\cdot,\cdot)$ is the MSE loss. 
We tried $\beta$’s in $\{0, 1, 10, 100, 1000\}$ and the $\beta$ in $[1, 100]$ achieves stable performance. 
Therefore, we empirically set $\beta=10$.
When solving (\ref{eq:emb}), we use Adam~\cite{kingma2014adam} with learning rate $0.001$ and set the number of epochs to $5$, and the size of batches as $100$. 
The proposed method based on cosine and RBF distances are denoted \textbf{GWL-C} and \textbf{GWL-R}, respectively.
Additionally, to highlight the benefit from joint graph matching and node-embedding learning, we consider a baseline that purely minimizes GW discrepancy based on data-driven distance matrices (denoted as \textbf{GWD}). 
The code is available on \url{https://github.com/HongtengXu/gwl}.

\subsection{Synthetic data}
We verify the feasibility of our GWL method by first considering two kinds of synthetic datasets.
The graphs in the first dataset imitate K-NN graphs with certain randomness, which is common in practical data science. 
The graphs in the second dataset yield to the Barab{\'a}si-Albert (BA) model~\cite{barabasi2016network}, which matches the statistics of real-world networks well.  
For the K-NN graph dataset, we simulate the source graph $G(\mathcal{V}_s,\mathcal{E}_s)$ as follows: for each $v_i\in\mathcal{V}_s$, we select $K\sim \text{Poisson}(0.1\times |\mathcal{V}_s|)$ nodes randomly from $\mathcal{V}_s\setminus v_i$, denoted as $\{v_i^k\}_{k=1}^{K}$. 
For each selected edge $(v_i, v_i^k)$, there are $w\sim \text{Poisson}(10)$ interactions between these two nodes. 
Accordingly, $\mathcal{E}_s$ is the union of all simulated $\{(v_i, v_i^k, w)\}_{i,k}$. 
The target graph $G(\mathcal{V}_t,\mathcal{E}_t)$ is constructed by first adding $q\%$ noisy nodes to the source graph, $i.e.$, $|\mathcal{V}_t|=(1+q\%)|\mathcal{V}_s|$, and then generating $q\%$ noisy edges between the nodes in $\mathcal{V}_t$ via the simulation method mentioned above, $i.e.$, $|\mathcal{E}_t|=(1+q\%)|\mathcal{E}_s|$.
Similarly, for the BA graph dataset, we first simulate the source graph with $|\mathcal{V}_s|$ nodes, and then simulate the target graph via adding  $q\%|\mathcal{V}_s|$ more nodes and corresponding edges.

\begin{figure}[t]
    \centering
    \includegraphics[width=1\linewidth]{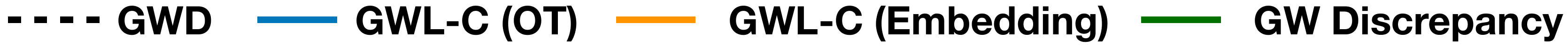}
    \subfigure[K-NN: $|\mathcal{V}_s|=50$]{
    \includegraphics[height=3.4cm]{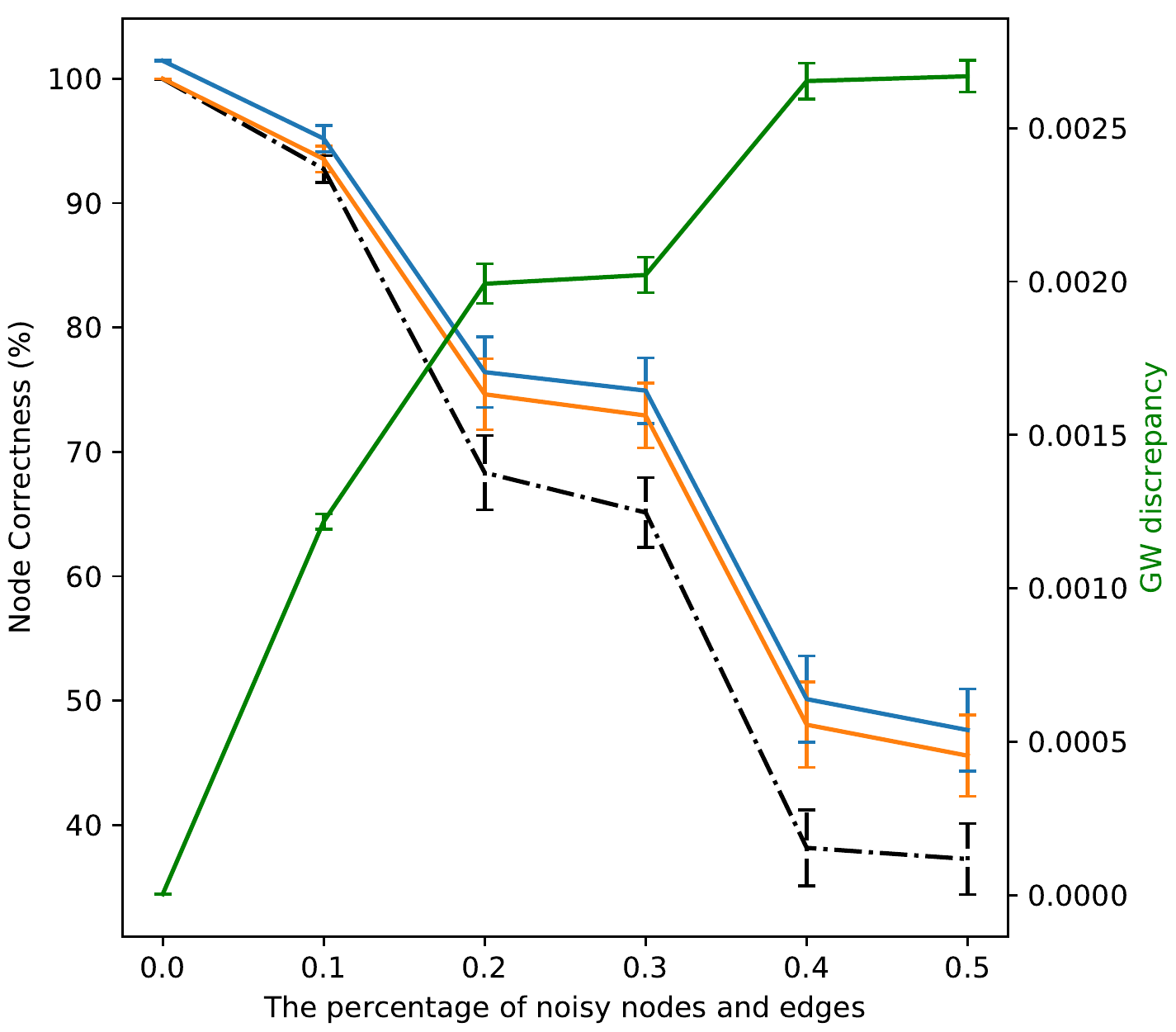}\label{fig:50}
    }
    \subfigure[BA: $|\mathcal{V}_s|=50$]{
    \includegraphics[height=3.4cm]{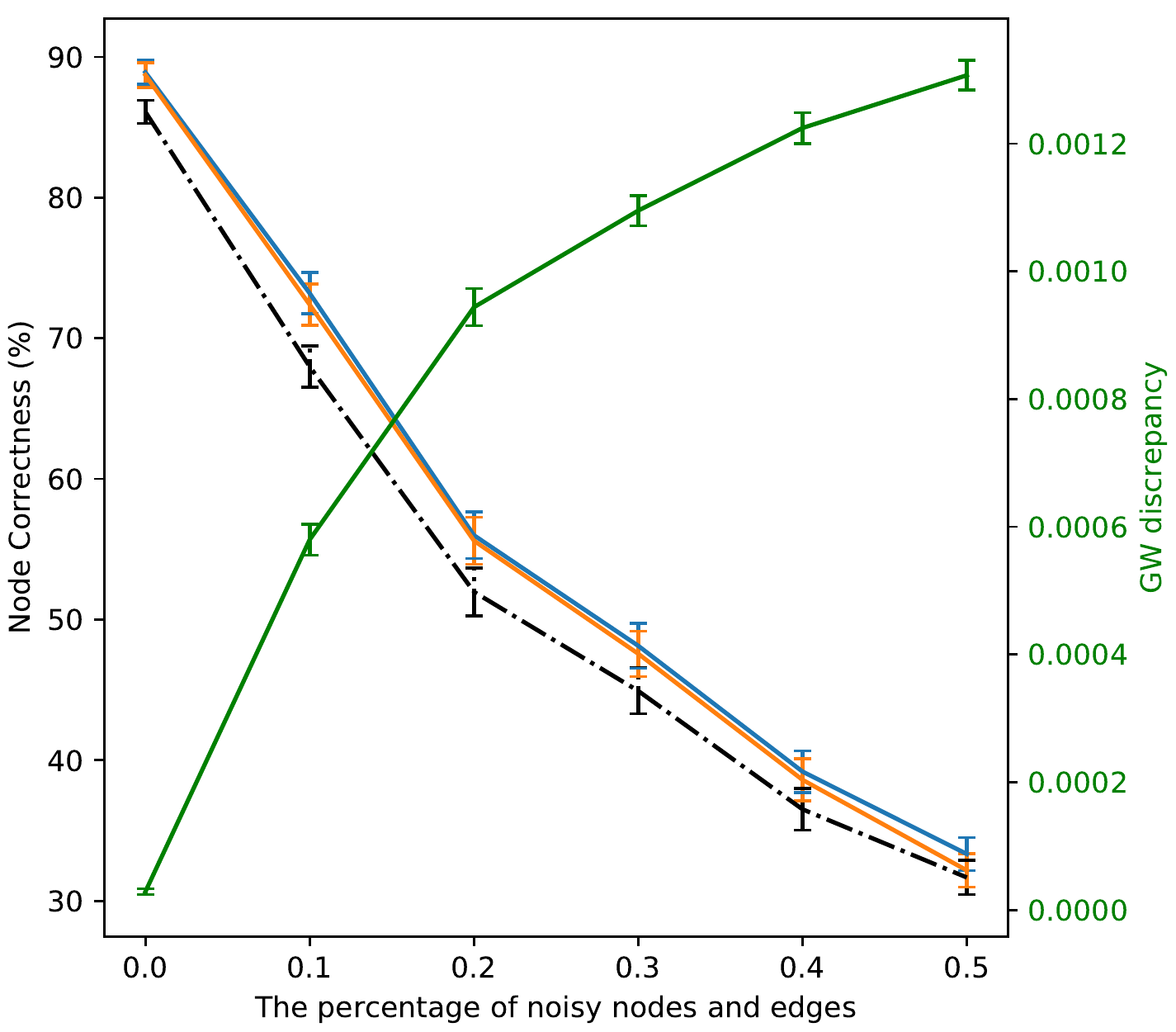}\label{fig:ba50}
    }
    \vspace{-10pt}
    \caption{The performance of our method on synthetic data.}\vspace{-10pt}
    \label{fig:syn}
\end{figure}

We set $|\mathcal{V}_s|\in\{50, 100\}$ and $q\in\{0, 10, 20, 30, 40, 50\}$. 
For each configuration, we simulate the source graph and the target one in $100$ trials. 
For each trial, we apply our method (and its baseline GWD) to match the graphs and calculate \emph{node correctness} as our measurement: Given the learned correspondence set $\mathcal{P}$ and the ground truth set of correspondences $\mathcal{P}_{real}$, we calculate percent node correctness as $NC=\frac{|\mathcal{P}\cap\mathcal{P}_{real}|}{|\mathcal{P}|}\times 100\%$. 
To analyze the rationality of the learned node embeddings, we construct $\mathcal{P}$ in two ways: for each $v_i^s\in \mathcal{V}_s$, we find its matched node $v_j^t\in\mathcal{V}_t$ via ($i$) $\arg\max_j \widehat{T}_{ij}$ (as shown in line 13 of Algorithm~\ref{alg2}) or ($ii$) $\arg\min_j \kappa(\bm{x}_i^s,\bm{x}_j^t)$. 
Additionally, the corresponding GW discrepancy is calculated as well. 
Assuming that the results in different trials are Gaussian distributed, we calculate the $95\%$ confidence interval for each measurement.

Figure~\ref{fig:syn} visualizes the performance of our GWL-C method and its baseline GWD when $|\mathcal{V}_s|=50$. 
More results are in the Supplementary Material.
When the target graph is identical to the source one ($i.e.$, $q=0$), the proposed Gromov-Wasserstein learning framework can achieve almost $100\%$ node correctness, and the GW discrepancy approaches zero. 
With the increase of $q$, the noise in the target graph becomes serious, and the GW discrepancy increases accordingly. 
It means that the GW discrepancy reflects the dissimilarity between the graphs indeed. 
Although the GWD is comparable to our GWL-C in the case with low noise level, it becomes much worse when $q>20$. 
This phenomenon supports our claim that learning node embeddings can improve the robustness of graph matching. 
Moreover, we find that the node correctness based on the optimal transport (blue curves) and that based on the embeddings (orange curves) are almost the same. 
This demonstrates that the embeddings of different graphs are on the same manifold, and their distances indicate the correspondences between graphs.


\subsection{MC3: Matching communication networks}
MC3 is a dataset used in the Mini-Challenge 3 of VAST Challenge 2018, which records the communication behavior among a company's employees on different networks.\footnote{\url{http://vacommunity.org/VAST+Challenge+2018+MC3}} 
The communications are categorized into two types: phone calls and emails between employees. 
According to the types of the communications, we obtain two networks, denoted as \emph{CallNet} and \emph{EmailNet}. 
Because an employee has two independent accounts in these two networks, we aim to link the accounts belonging to the same employee. 
We test our method on a subset of the MC3 dataset, which contains $622$ employees and their communications through phone calls and emails. 
In this subset, for each selected employee there is at least one employee in a network (either CallNet or EmailNet) having over $10$ times communications with him/her, which ensures that each node has at least one reliable edge. 
Additionally, for each network, we can control the density of its edge by thresholding the count of interactions. When we only keep the edges corresponding to the communications happening more than $8$ times, we obtain two sparse graphs: the CallNet contains $1,228$ edges and the EmailNet contains $1,235$ edges. 
When we keep all the communications and the corresponding edges, we obtain two dense graphs, the CallNet contains $141,846$ edges and the EmailNet contains $115,782$ edges. 
Generally, experience indicates that matching dense graphs is much more difficult than matching sparse ones.

We compare our methods (GWL-R and GWL-C) with well-known graph matching methods:
the graduated assignment algorithm (GAA)~\cite{gold1996graduated}, the low-rank spectral alignment (LRSA)~\cite{nassar2018low}, TAME~\cite{mohammadi2017triangular}, GRAAL\footnote{\url{http://www0.cs.ucl.ac.uk/staff/natasa/GRAAL}.}, MI-GRAAL\footnote{\url{http://www0.cs.ucl.ac.uk/staff/natasa/MI-GRAAL}.}, MAGNA++\footnote{\url{https://www3.nd.edu/~cone/MAGNA++}.}, HugAlign and NETAL.\footnote{\url{http://ttic.uchicago.edu/~hashemifar}.}
These alternatives achieve the state-of-the-art performance on matching large-scale graphs, $e.g.$, protein networks.
Table~\ref{tab:mc3} lists the matching results obtained by the different methods.\footnote{For GWD, GWL-R and GWL-C, here we show the node correctness calculated based on the learned optimal transport.} 
For the alternative methods, their best results in $10$ trials are listed. 
We can find that their performance on sparse and dense graphs is inconsistent.
For example, GRAAL works almost as well as our GWL-R and GWL-C for sparse graphs, but its matching result becomes much worse for dense graphs.
For the baseline GWD, it is inferior to most graph-matching methods on node correctness, because it purely minimizes the GW discrepancy based on the information of pairwise interactions ($i.e.$, edges). 
Additionally, GWD merely relies on data-driven distance matrices, which is sensitive to the noise in the graphs. 
However, when we take node embeddings (with dimension $D=100$) into account, the proposed GWL-R and GWL-C outperform GWD and other considered approaches consistently, on both sparse and dense graphs.

\begin{figure}[t]
    \centering
    \subfigure[Stability and convergence]{
    \includegraphics[height=3.4cm]{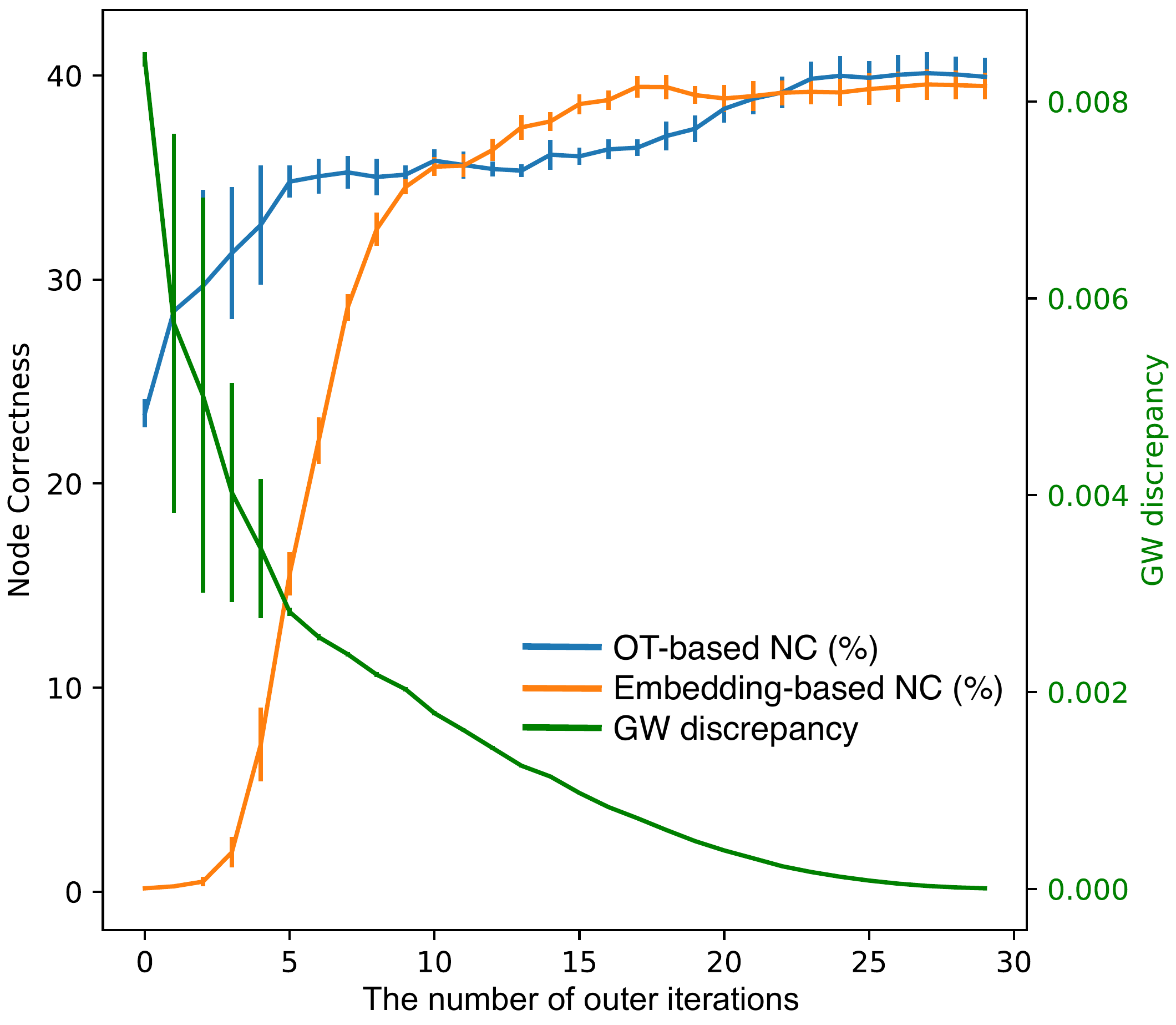}\label{fig:conv}
    }
    \subfigure[Learned embeddings]{
    \includegraphics[height=3.4cm]{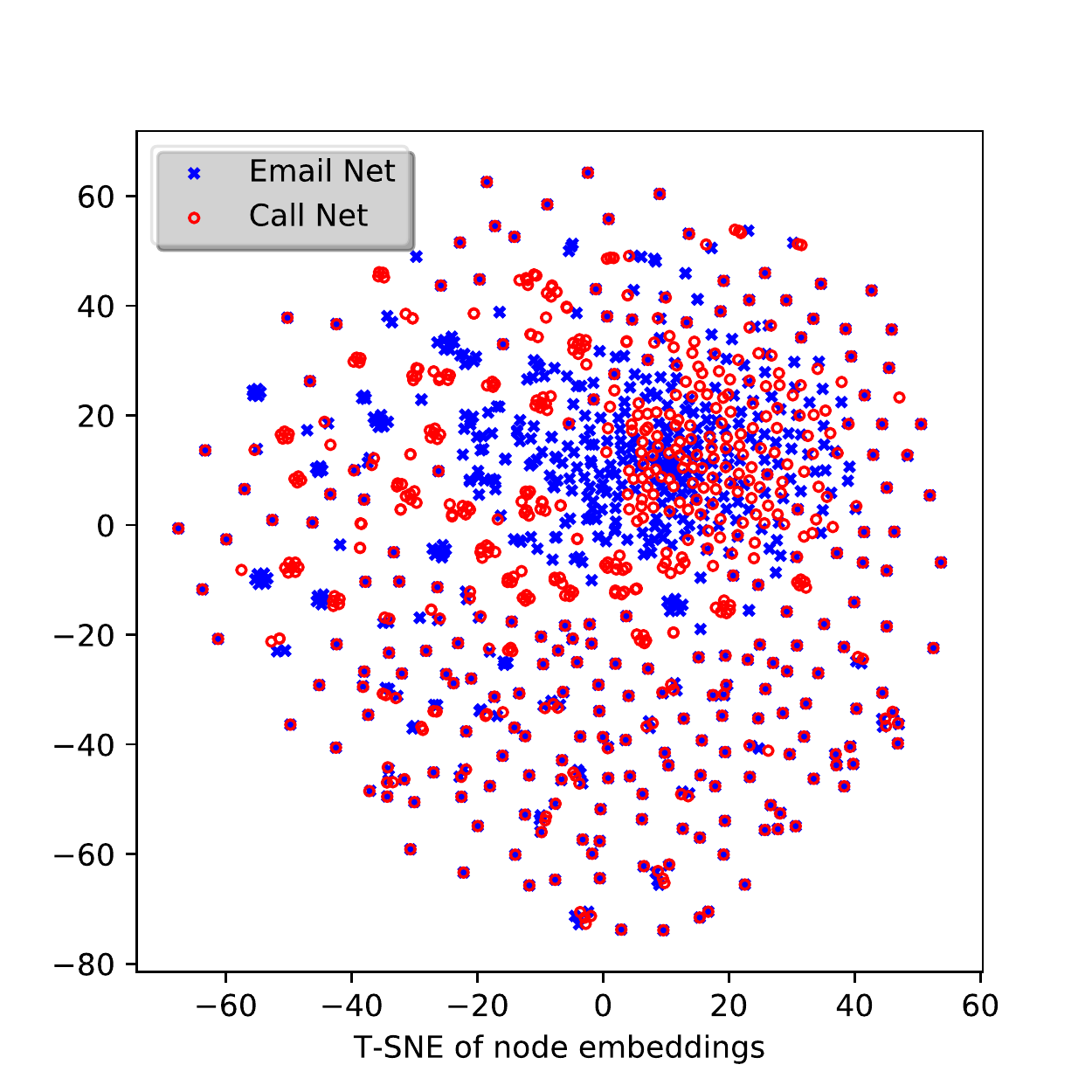}\label{fig:epoch29}
    }
    \vspace{-10pt}
    \caption{Visualization of typical experimental results.}\vspace{-10pt}
    \label{fig:typical}
\end{figure}

\begin{table}[!t]
\caption{Communication network matching results.}
\centering
\small{
\setlength{\tabcolsep}{5pt}
\begin{tabular}{c|c|c
} 
\hline\hline
\multirow{2}{*}{Method} 
&{Call$\rightarrow$Email (Sparse)}
&{Call$\rightarrow$Email (Dense)}\\
\cline{2-3}
         &Node Correctness (\%) &Node Correctness (\%) \\
\hline
GAA      &34.22   &0.53\\
LRSA     &38.20   &2.93\\
TAME     &37.39   &2.67\\
GRAAL    &39.67   &0.48\\
MI-GRAAL &35.53   &0.64\\
MAGNA++  &7.88    &0.09\\
HugAlign &36.21   &3.86\\
NETAL    &36.87   &1.77\\
\hline
GWD      &23.16$\pm$0.46   &1.77$\pm$0.22\\
GWL-R    &39.64$\pm$0.57   &3.80$\pm$0.23\\
GWL-C    &\textbf{40.45}$\pm$0.53   &\textbf{4.23}$\pm$0.27\\
\hline\hline
\end{tabular}\label{tab:mc3}
}\vspace{-10pt}
\end{table}

To demonstrate the convergence and the stability of our method, we run GWD, GWL-R and GWL-C in $10$ trials with different initialization. 
For each method, its node correctness is calculated based on optimal transport and the embedding-based distance matrix. 
The $95\%$-confidence interval of the node correctness is estimated as well, as shown in Table~\ref{tab:mc3}.
We find that the proposed method has good stability and outperforms other methods with high confidence. 
Figure~\ref{fig:conv} visualizes the GW discrepancy and the node correctness with respect to the number of outer iterations; the $95\%$-confidence intervals are shown as well. 
In Figure~\ref{fig:conv}, we find that the GW discrepancy decreases and the two kinds of node correctness increase accordingly and become consistent with the increase of iterations, which means that the embeddings we learn and their distances indeed reflect the correspondence between the two graphs. 
Figure~\ref{fig:epoch29} visualizes the learned embeddings with the help of t-SNE~\cite{maaten2008visualizing}. 
We find that the learned node embeddings of different graphs are on the same manifold and the overlapped embeddings indicate matched pairs.

\subsection{MIMIC-III: Procedure recommendation}
Besides typical graph matching, our method has potential for other applications, like recommendation systems. 
Such systems recommend items to users according to the distance/similarity between their embeddings. 
Traditional methods~\cite{rendle2009bpr,chen2018sequential} learn the embeddings of users and items purely based on their interactions. 
Recent work~\cite{monti2017geometric,ying2018graph} shows that considering the user network and/or item network is beneficial to improve recommendation results. 
Such a strategy is also applicable to our Gromov-Wasserstein learning framework: given the network of users, the network of items, and the observed interactions between them ($i.e.$, partial correspondences between the graphs), we learn the embeddings of users and items and the optimal transport between them via minimizing the GW discrepancy between the networks. 
Because the learned embeddings are on the same manifold, we can calculate the distance between a user and an item directly via the cosine-based distance or the RBF-based distance.
Accordingly, we recommend each user with the items with shortest distances. 
For our method, the only difference between the recommendation task and previous graph matching task is that we observed some interactions, $i.e.$, the $w_{ij}$ between source node $v_i\in\mathcal{V}_s$ and target node $v_j\in\mathcal{V}_t$. 
In such a situation, we take the optional regularizer in~(\ref{eq:reg}) into account.
Based on observed $w_{ij}$'s, the elements of the $\bm{C}_{st}$ in (\ref{eq:reg}) are calculated via (\ref{eq:zipf}).

We test the feasibility of our method on the MIMIC-III dataset~\cite{johnson2016mimic}, which contains patient admissions in a hospital.
Each admission is represented as a sequence of ICD (International Classification of Diseases) codes of the diseases and the procedures. 
The diseases (procedures) appearing in the same admission construct the interactions of the disease (procedure) graph.
We aim to recommend suitable procedures for patients, according to their disease characteristics. 
To achieve this, we learn the embeddings of the ICD codes for the diseases and the procedures with the help of various methods, and measure the distance between the embeddings. 
We compare the proposed GWL method with the following baselines: 
$i$) treating the admission sequences as sentences and learning the embeddings of ICD codes via traditional word embedding methods like \textbf{Word2Vec}~\cite{mikolov2013efficient} and \textbf{GloVe}~\cite{pennington2014glove}; 
$ii$) the distilled Wasserstein learning (\textbf{DWL}) method in~\cite{xu2018distilled}, which trains the embeddings from scratch or fine-tunes Word2Vec's embeddings based on a Wasserstein topic model; and  
$iii$) the GWD method that minimizes the GW discrepancy purely based on the data-driven distance matrices, and then learns the embeddings regularized by the learned optimal transport. 
The GWD method is equivalent to applying our GWL method and setting the number of outer iterations $M=1$.
For the GWD method, we also consider the cosine- and RBF-based distances when learning embeddings, denoted as \textbf{GWD-C} and \textbf{GWD-R}, respectively.

For fairness of comparison, we use a subset of the MIMIC-III dataset provided by~\cite{xu2018distilled}, which contains $11,086$ patient admissions, corresponding to $56$ diseases and $25$ procedures. 
For all the methods, we use $50\%$ of the admissions for training, $25\%$ for validation, and the remaining $25\%$ for testing. 
In the testing phase, for the $i$-th admission, $i=1,...,I$, we may recommend a list of procedures with length $L$, denoted as $E_i$, based on its diseases and evaluate recommendation results based on the ground truth list of procedures, denoted as $T_i$. 
Given $\{E_i, T_i\}$, we calculate the top-$L$ precision, recall and F1-score as follows:
$P=\sum_{i=1}^I P_i=\sum_{i=1}^{I}\frac{|E_i\cap T_i|}{|E_i|}$,
$R=\sum_{i=1}^I R_i=\sum_{i=1}^{I}\frac{|E_i\cap T_i|}{|T_i|}$,
$F1=\sum_{i=1}^I\frac{2P_i R_i}{P_i+R_i}$. 
Table~\ref{tab:mimic} shows the results of various methods with $L=1$ and $5$.
We find that our GWL method outperforms the alternatives, especially on the top-1 measurements.

We analyze the learned optimal transport between diseases and procedures from a clinical viewpoint. 
In particular, we normalize the transport matrix, ensuring its maximum value is $1$.
For each disease, we find the corresponding procedures $i$) with the maximum optimal transports and $ii$) with $\widehat{T}_{ij}>0.15$. 
We asked two clinical researchers to check the pairs we find; 
they confirmed that for over $77.42\%$ of the pairs, either the procedures are clearly related to the treatments of the diseases, or the procedures clearly lead to the diseases as side effects or complications (other relationships may be less clear, but are implied by the data), $e.g.$, ``(dV3001) Single liveborn, born in hospital, delivered by cesarean section $\leftrightarrow$ (p640) Circumcision''.
The learned optimal transport, and all pairs of ICD codes and their evaluation results are shown in the Supplementary Material.

\begin{table}[t]
\caption{Top-$N$ procedure recommendation results.}
\centering
\small{
\setlength{\tabcolsep}{5pt}
\begin{tabular}{c|c@{\hspace{6pt}}c@{\hspace{6pt}}c|c@{\hspace{6pt}}c@{\hspace{6pt}}c} 
\hline\hline
\multirow{2}{*}{Method} 
&\multicolumn{3}{c|}{Top-1 (\%)} 
&\multicolumn{3}{c}{Top-5 (\%)}\\
\cline{2-7}
&P &R &F1
&P &R &F1\\
\hline
Word2Vec
&39.95 &13.27 &18.25
&28.89 &46.98 &32.59\\
GloVe  	
&32.66 &13.01 &17.22
&27.93 &44.79 &31.47\\
DWL (Scratch)   	
&37.89 &12.42 &17.16
&27.39 &43.81 &30.81\\
DWL (Finetune)  	
&40.00 &13.76 &18.71					
&30.59 &\textbf{48.56} &34.28\\
\hline
GWD-R
&46.29 &17.01 &22.32
&31.82 &43.81 &33.77\\
GWD-C
&43.16 &15.79 &20.77
&31.42 &42.99 &33.25\\
GWL-R
&46.20 &16.93 &22.22
&32.03 &44.75 &34.18\\
GWL-C
&\textbf{47.46} &\textbf{17.25} &\textbf{22.71}
&\textbf{32.09} &45.64 &\textbf{34.31}\\
\hline\hline
\end{tabular}\label{tab:mimic}
}\vspace{-10pt}
\end{table}

\section{Conclusions and Future Work}
We have proposed a Gromov-Wasserstein learning method to unify graph matching and the learning of node embeddings into a single framework. 
We show that such joint learning is beneficial to each of the objectives, obtaining superior performance in various matching tasks.
In the future, we plan to extend our method to multi-graph matching~\cite{yan2016multi}, which may be related to Gromov-Wasserstein barycenter~\cite{peyre2016gromov} and its learning method. 
Additionally, to improve the scalability of our method, we will explore new Gromov-Wasserstein learning algorithms.

\textbf{Acknowledgments} This research was supported in part by DARPA, DOE, NIH, ONR and NSF. We thank Dr. Matthew Engelhard and Rachel Draelos for evaluating our results. 
We also thank Wenlin Wang for helpful discussions.

\newpage
\bibliographystyle{icml2019}
\bibliography{rgwl}

\newpage
\section{Supplementary Material}
\subsection{The scheme of proposed proximal point method}
In the $m$-th ouer iteration, we learn the optimal transport iteratively. 
Particularly, in the $n$-th inner iteration, we update the target optimal transport via solving (\ref{eq:ipot}) based on the Sinkhorn-Knopp algorithm~\cite{sinkhorn1967concerning,cuturi2013sinkhorn}. 
Algorithm~\ref{alg1} gives the details of our proximal point method in the $m$-th outer iteration, where $\text{diag}(\cdot)$ converts a vector to a diagonal matrix, and $\odot$ and $\frac{\cdot}{\cdot}$ represent element-wise multiplication and division, respectively.

\subsection{The convergence of each updating step}
The proposed proximal point method decomposes a nonconvex optimization problem into a series of convex updating steps.
Each updating step corresponds to the solution to a regularized optimal transport problem, which is solved via $J$ Sinkhorn projections. 
The work in~\cite{altschuler2017near} proves that solving the regularized optimal transport based on Sinkhorn projections is with linear convergence. 
The work in~\cite{xie2018fast} further proves that the linear convergence holds even just applying one-step Sinkhorn projection in each updating step ($i.e.$, $J=1$). 
Therefore, the updating steps of the proposed method are with linear convergence.

\subsection{Global convergence: The proof of Proposition~\ref{prop}}
\textbf{Proposition 3.1} \emph{Every limit point generated by our proximal point method, $i.e.$, $\lim_{n\rightarrow\infty}\bm{T}^{(n)}$, is a stationary point of the problem (\ref{eq:ot}).}
\begin{proof}
When learning the target optimal transport, the original optimization problem (\ref{eq:ot}) is with a nonconvex and differentiable objective function 
\begin{eqnarray}
\begin{aligned}
f(\bm{T})=&\bm{L}(\bm{C}_s(\bm{X}_s^{(m)}), \bm{C}_t(\bm{X}_t^{(m)}),\bm{T}),\bm{T} \rangle\\
&+\alpha\langle \bm{K}(\bm{X}_s^{(m)}, \bm{X}_t^{(m)}),\bm{T} \rangle,
\end{aligned}
\end{eqnarray}
and a closed convex set $\mathcal{T}=\Pi(\bm{\mu}_s,\bm{\mu}_t)$ as the constraint of $\bm{T}$. As a special case of successive upper-bound minimization (SUM), our proximal point method solves (\ref{eq:ot}) via optimizing a sequence of approximate objective functions: starting from a feasible point $\bm{T}^{(0)}$, the algorithm generates a sequence $\{\bm{T}^{(n)}\}$ according to the update rule:
\begin{eqnarray}
\begin{aligned}
\bm{T}^{(n+1)} = \arg\sideset{}{_{\bm{T}\in \mathcal{T}}}\min u(\bm{T}, \bm{T}^{(n)}),
\end{aligned}
\end{eqnarray}
where 
\begin{eqnarray}
\begin{aligned}
u(\bm{T},\bm{T}^{(n)})=f(\bm{T}) + \gamma \mbox{KL}(\bm{T}\lVert \bm{T}^{(n)}).
\end{aligned}
\end{eqnarray} 
is an approximation of $f(\bm{T})$ at the $n$-th iteration, and $\bm{T}^{(n)}$ is the point generated in the previous iteration.

Obviously, we have
\begin{itemize}
    \item[C1:] $u(\bm{T}, \bm{T}')$ is continuous in $(\bm{T}, \bm{T}')$.
\end{itemize}
Additionally, because $\mbox{KL}(\bm{T}\lVert\bm{T}')=\sum_{ij}T_{ij}\log\frac{T_{ij}}{T_{ij}'} -T_{ij} +T_{ij}'\geq 0$ and the equality holds only when $\bm{T}'=\bm{T}$, we have
\begin{itemize}
    \item[C2:] $u(\bm{T}, \bm{T})=f(\bm{T})$ $\forall~\bm{T}\in\mathcal{T}$.
    \item[C3:] $u(\bm{T}, \bm{T}')\geq f(\bm{T})$ $\forall~\bm{T},\bm{T}'\in \mathcal{T}$.
\end{itemize}
According to the Proposition 1 in~\cite{razaviyayn2013unified}, when the conditions C2 and C3 are satisfied, for the differentiable function $f(\bm{T})$ and its global upper bound $u(\bm{T},\bm{T}')$, we have
\begin{itemize}
    \item[C4:] $u'(\bm{T}, \bm{T}'; \bm{D})|_{\bm{T}=\bm{T}'}=f'(\bm{T}';\bm{D})$ $\forall~\bm{D}$ with $\bm{T}'+\bm{D}\in \mathcal{T}$, where 
    \begin{eqnarray*}
    \begin{aligned}
    f'(\bm{T};\bm{D}) :=\lim_{\delta\rightarrow 0}\inf \frac{f(\bm{T} +\delta \bm{D})-f(\bm{T})}{\delta}
    \end{aligned}
    \end{eqnarray*}
    is the directional derivative of $f(\bm{T})$ along the direction $\bm{D}$, and
    $u'(\bm{T}, \bm{T}'; d)$ is the directional derivative only with respect to $\bm{T}$.
\end{itemize}
According to the Theorem 1 in~\cite{razaviyayn2013unified}, when the approximate objective function $u(\cdot, \bm{T}^{(n)})$ in each iteration satisfies C1-C4, every limit point generated by the proposed method, $\lim_{n\rightarrow \infty}\bm{T}^{(n)}$, is a stationary point of the original problem (\ref{eq:ot}).
\end{proof}

\begin{algorithm}[t]
	\caption{Proximal Point Method for GW Discrepancy}
	\label{alg1}
	\begin{algorithmic}[1]
		\STATE \textbf{Input:} $\{\bm{C}_s,\bm{C}_t\}$, $\{\bm{\mu}_s,\bm{\mu}_t\}$, current embeddings $\{\bm{X}_s^{(m)},\bm{X}_t^{(m)}\}$, $\gamma$, the number of inner iterations $N$. 
		\STATE \textbf{Output:} $\widehat{\bm{T}}^{(m+1)}$.
		\STATE Initialize $\bm{T}^{(0)}=\bm{\mu}_s\bm{\mu}_t^{\top}$ and $\bm{a}=\bm{\mu}_s$.
		\FOR{$n=0:N-1$}
		\STATE Calculate the $\bm{C}^{(m, n)}$ in (\ref{eq:sinkhorn}).
		\STATE Set $\bm{G}=\exp(-\frac{\bm{C}^{(m,n)}}{\gamma})\odot \bm{T}^{(n)}$.
		\STATE $\backslash\backslash$ \texttt{Sinkhorn-Knopp algorithm:}
		\FOR{$j=1:J$}
        \STATE $\bm{b}=\frac{\bm{\mu}_{Y}}{\bm{G}^{\top}\bm{a}}$	
        \STATE $\bm{a}=\frac{\bm{\mu}_{X}}{\bm{G}\bm{b}}$
		\ENDFOR
		\STATE $\bm{T}^{(n+1)}=\text{diag}(\bm{a})\bm{G}\text{diag}(\bm{b})$.
		\ENDFOR
		\STATE $\widehat{\bm{T}}^{(m+1)}=\bm{T}^{(N)}$.
	\end{algorithmic}
\end{algorithm}

\begin{figure*}[!t]
    \centering
    \subfigure[$J=1$, $\gamma=1e-3$]{
    \includegraphics[width=3.7cm]{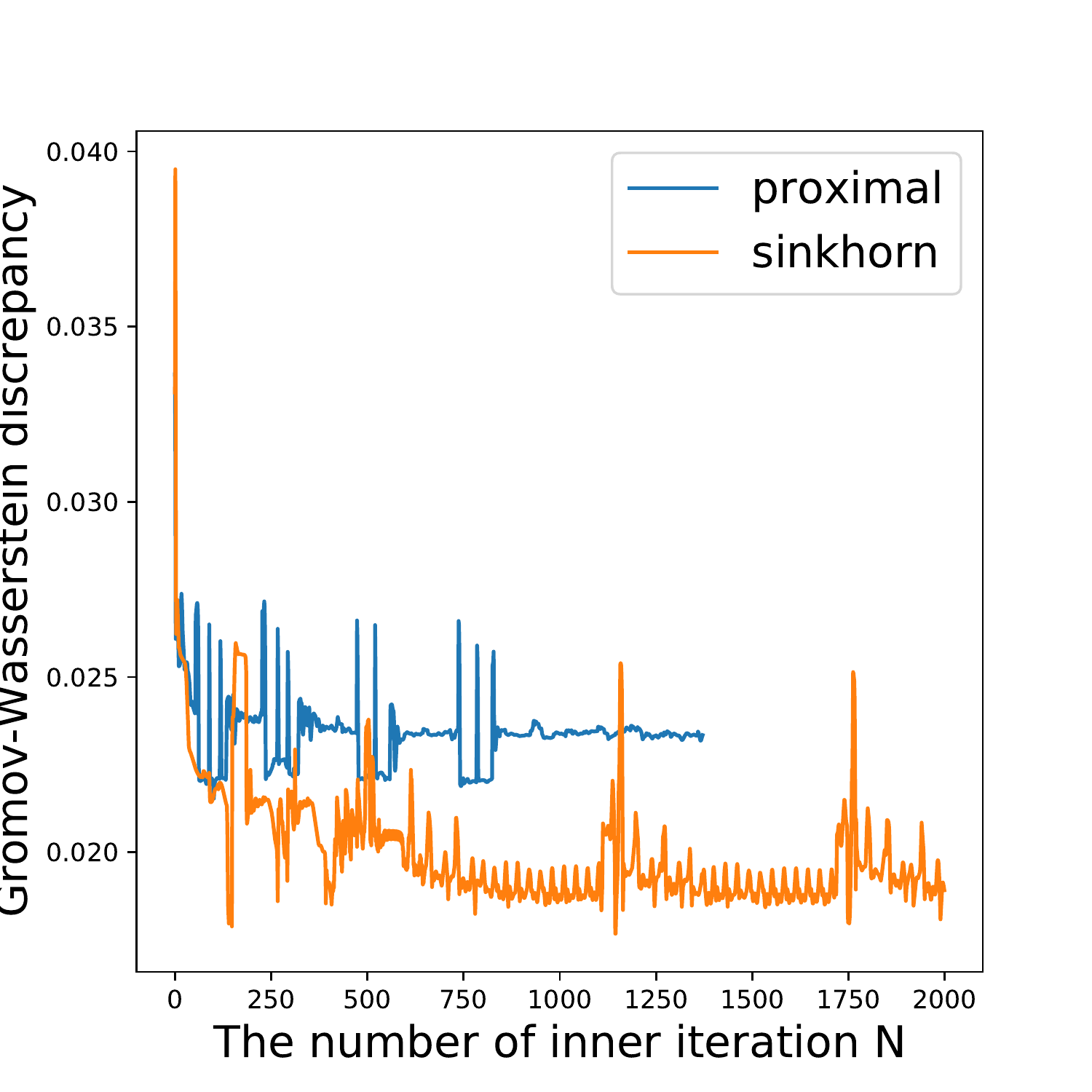}
    }
    \subfigure[$J=1$, $\gamma=1e-2$]{
    \includegraphics[width=3.7cm]{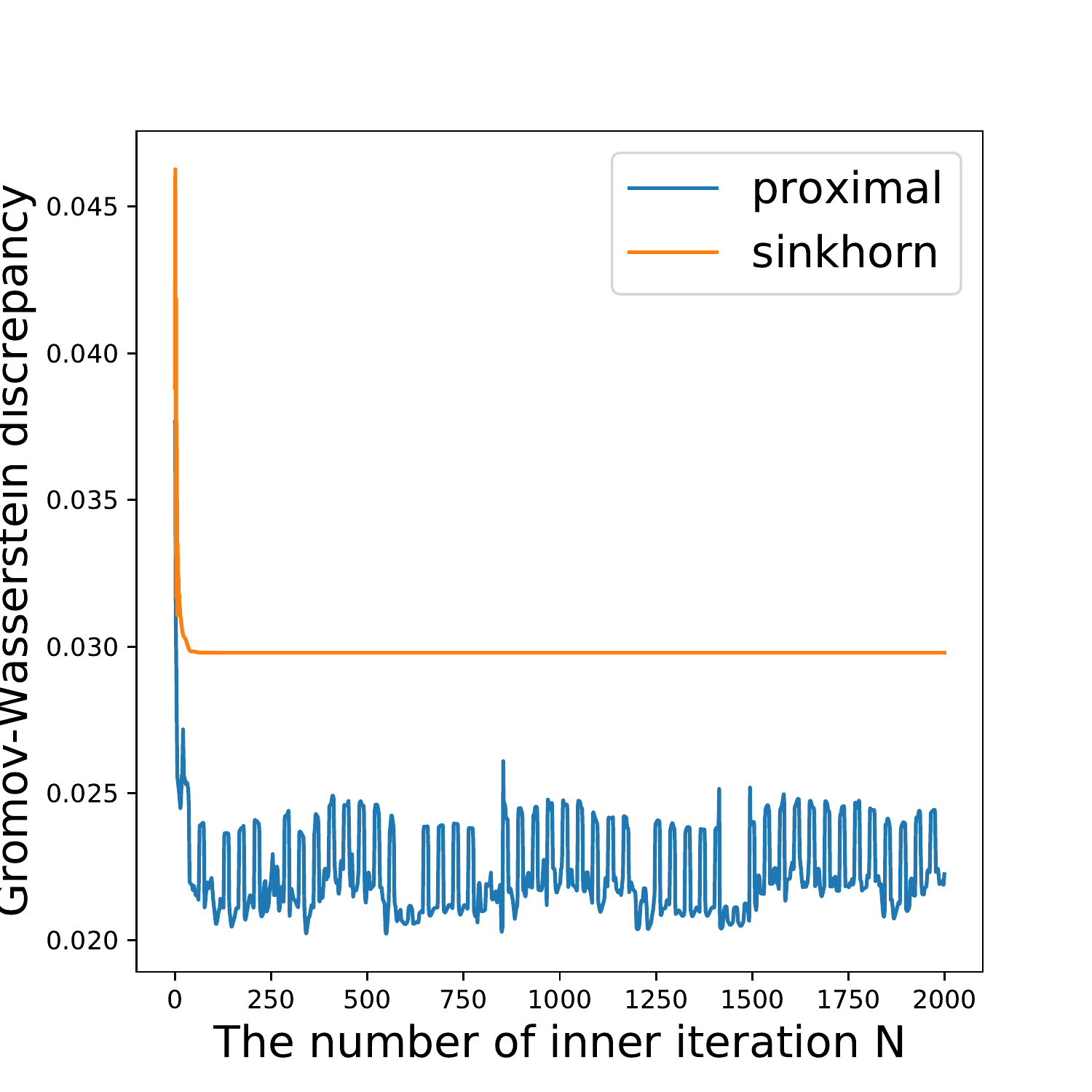}
    }
    \subfigure[$J=1$, $\gamma=1e-1$]{
    \includegraphics[width=3.7cm]{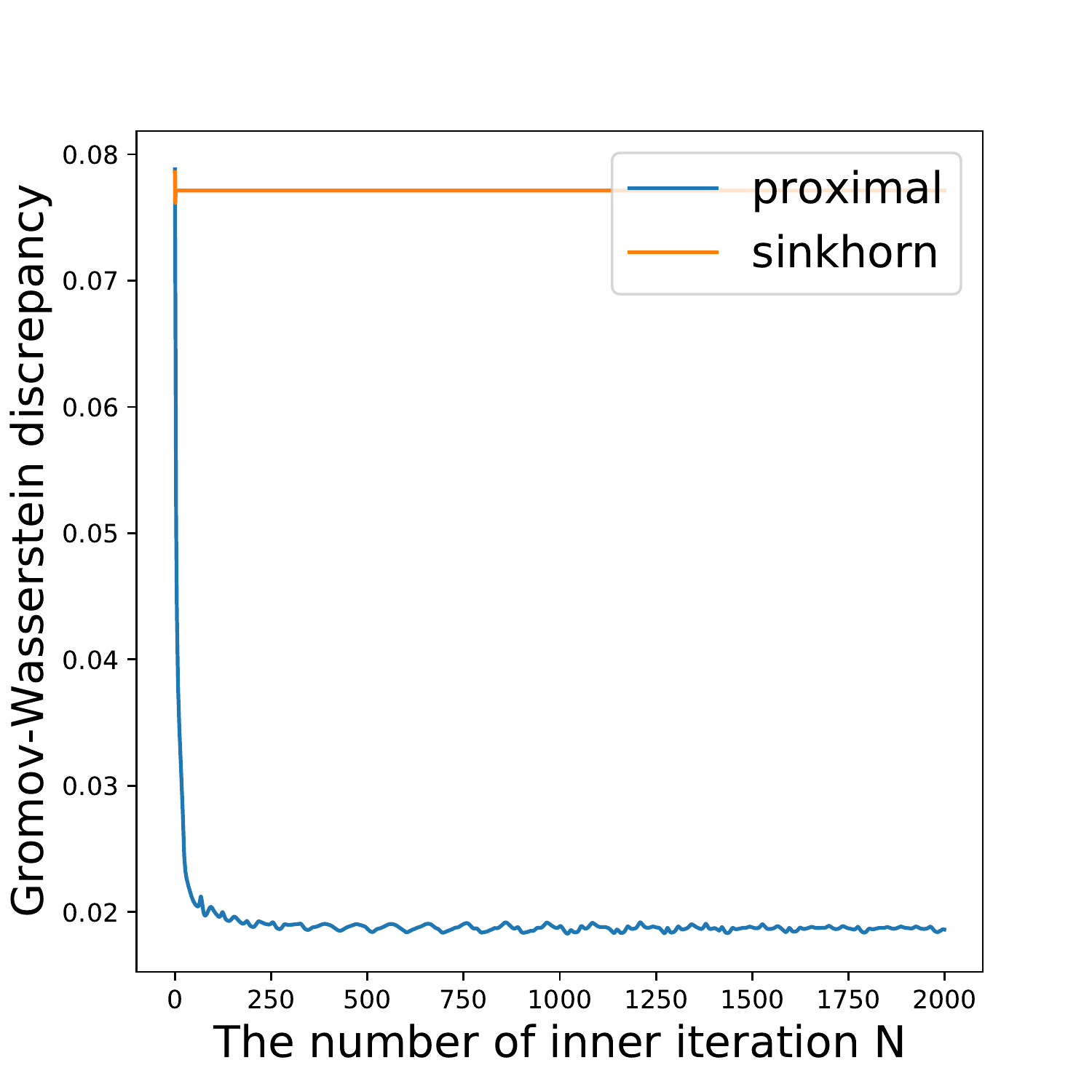}
    }
    \subfigure[$J=1$, $\gamma=1$]{
    \includegraphics[width=3.7cm]{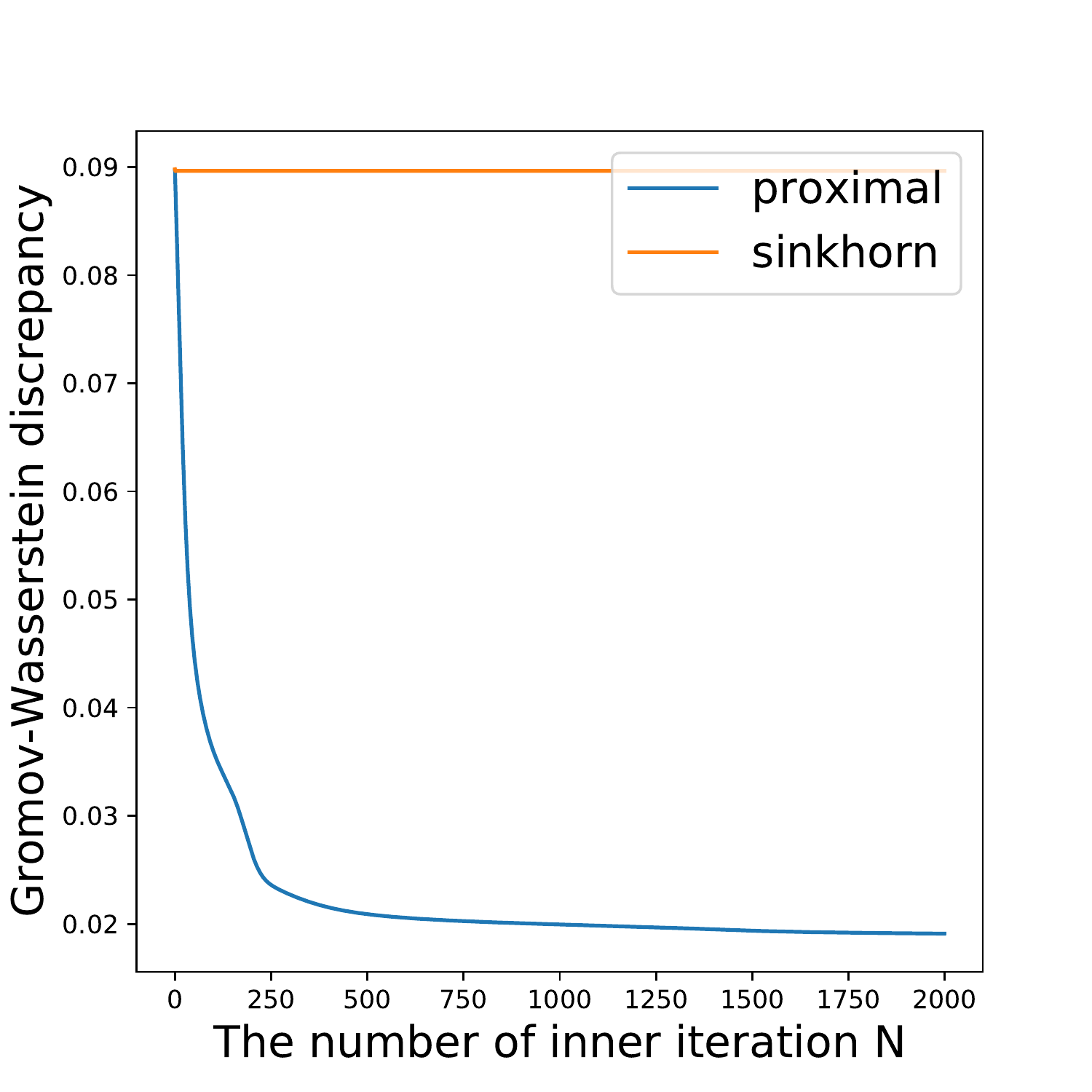}
    }
    \subfigure[$J=10$, $\gamma=1e-3$]{
    \includegraphics[width=3.7cm]{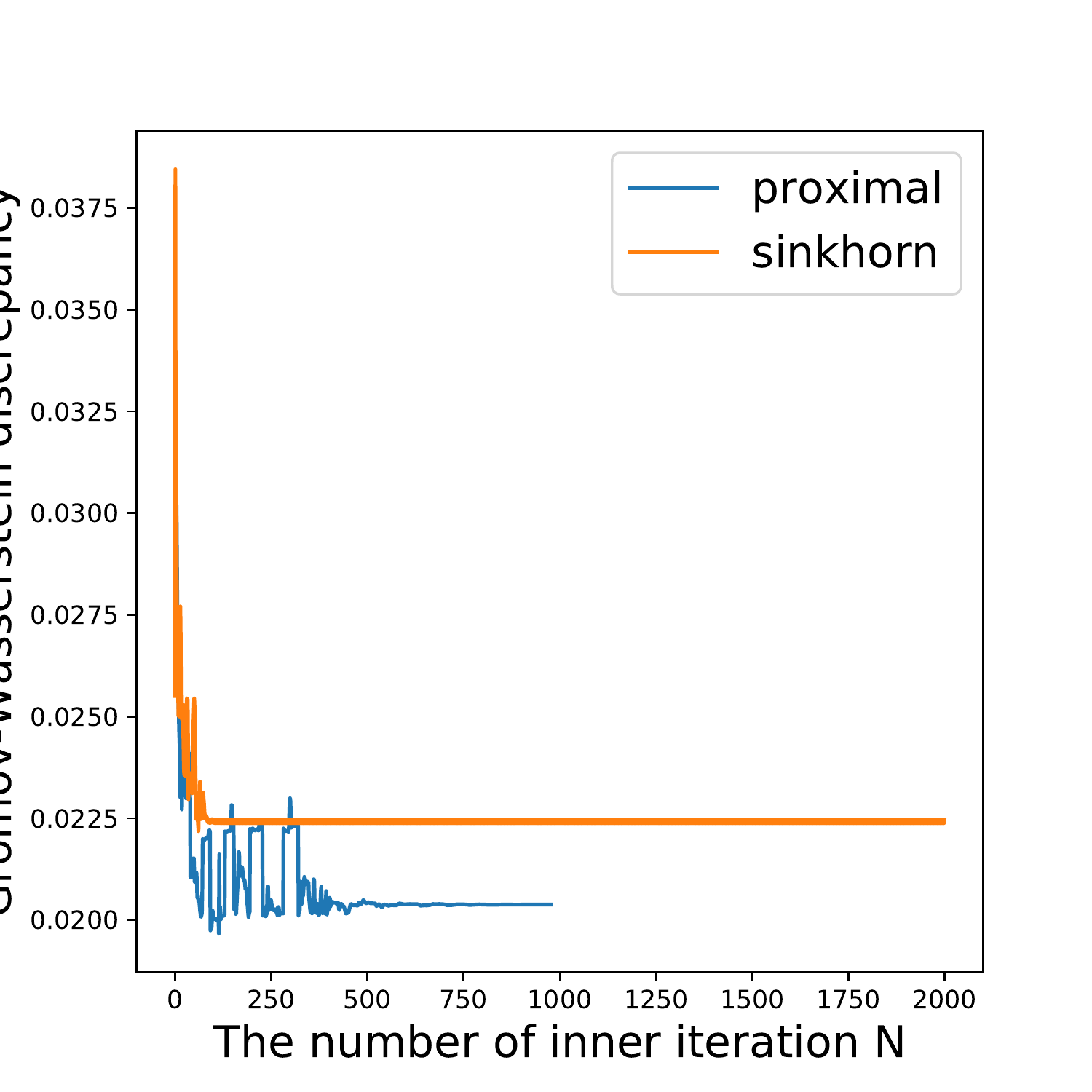}
    }
    \subfigure[$J=10$, $\gamma=1e-2$]{
    \includegraphics[width=3.7cm]{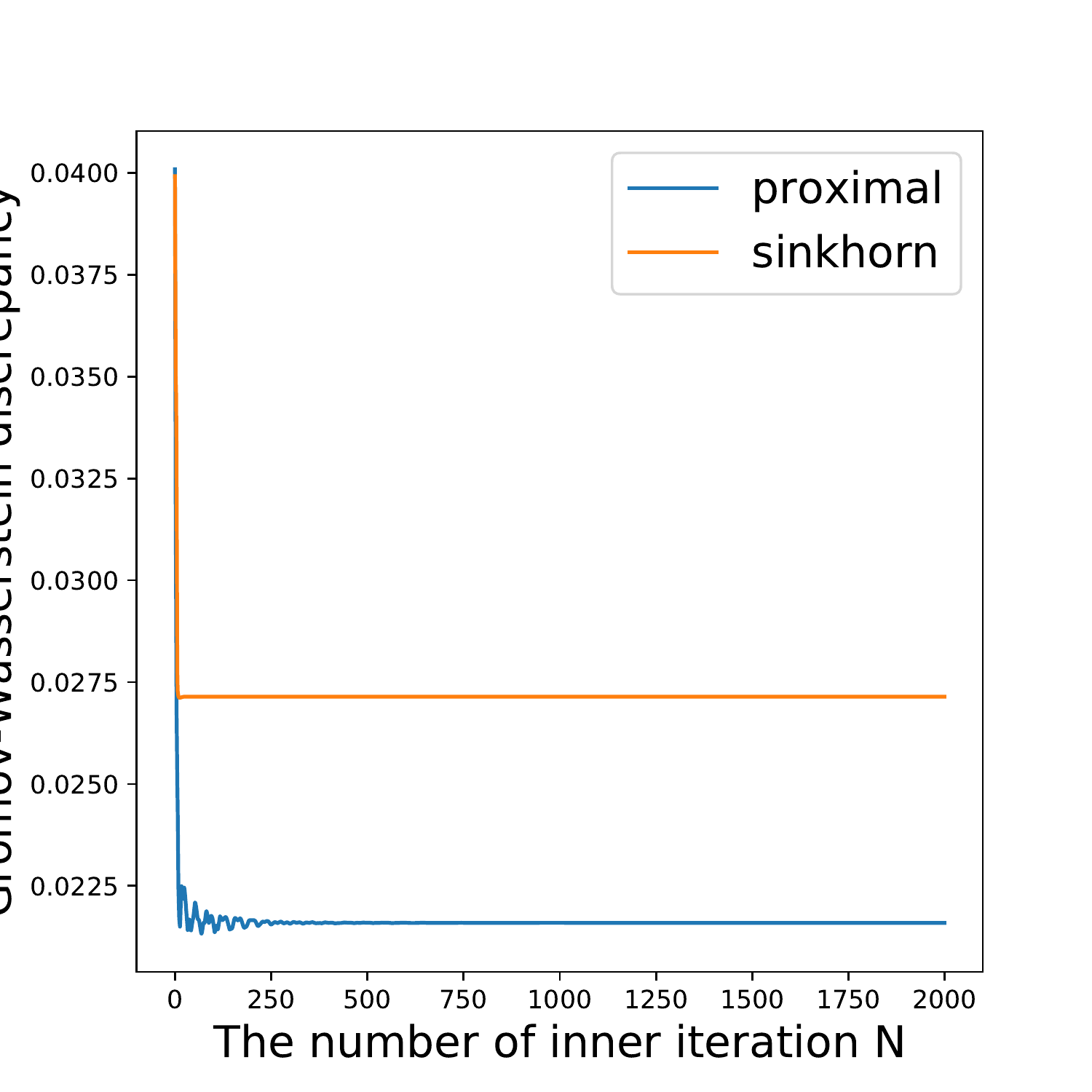}
    }
    \subfigure[$J=10$, $\gamma=1e-1$]{
    \includegraphics[width=3.7cm]{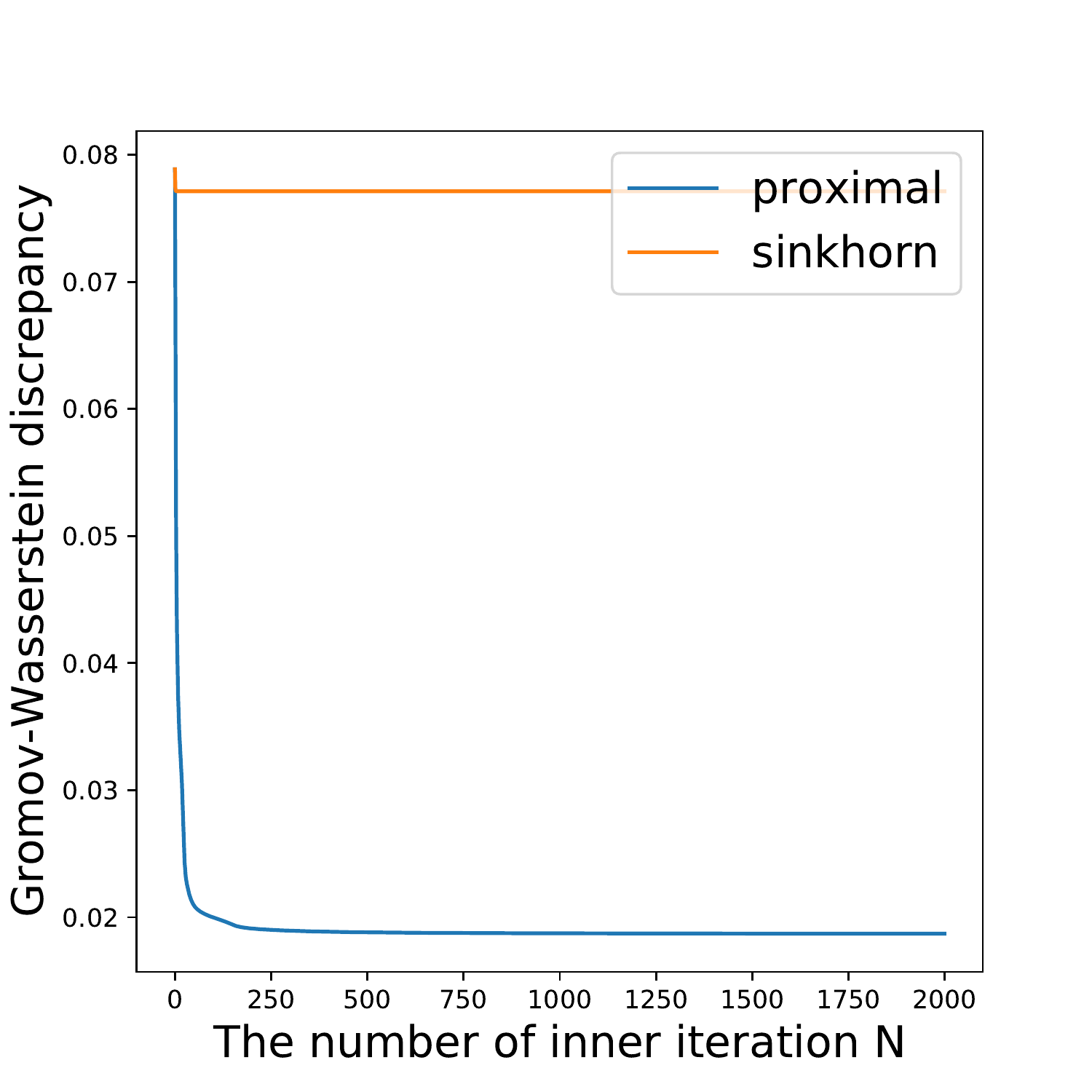}
    }
    \subfigure[$J=10$, $\gamma=1$]{
    \includegraphics[width=3.7cm]{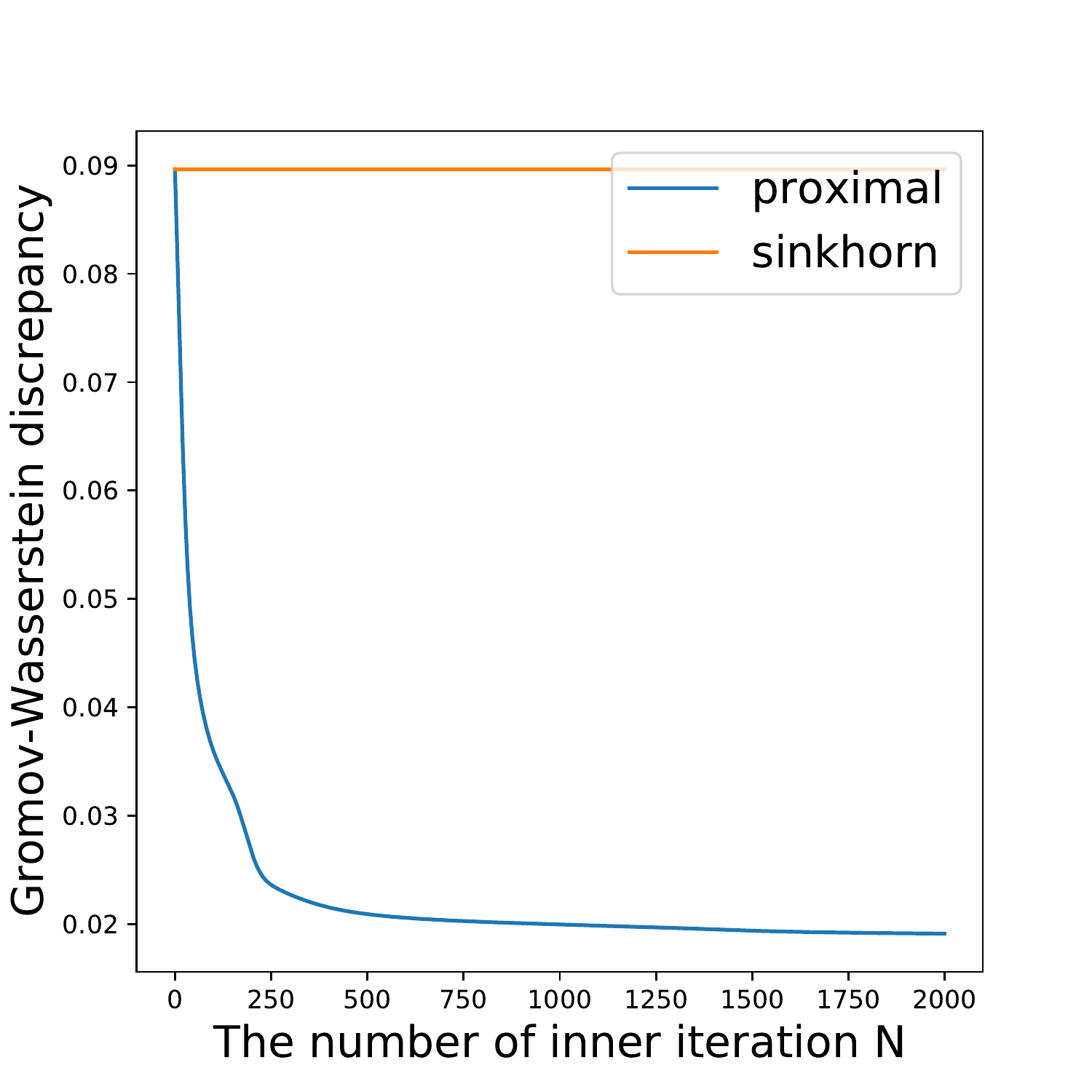}
    }
    \subfigure[$J=100$, $\gamma=1e-3$]{
    \includegraphics[width=3.7cm]{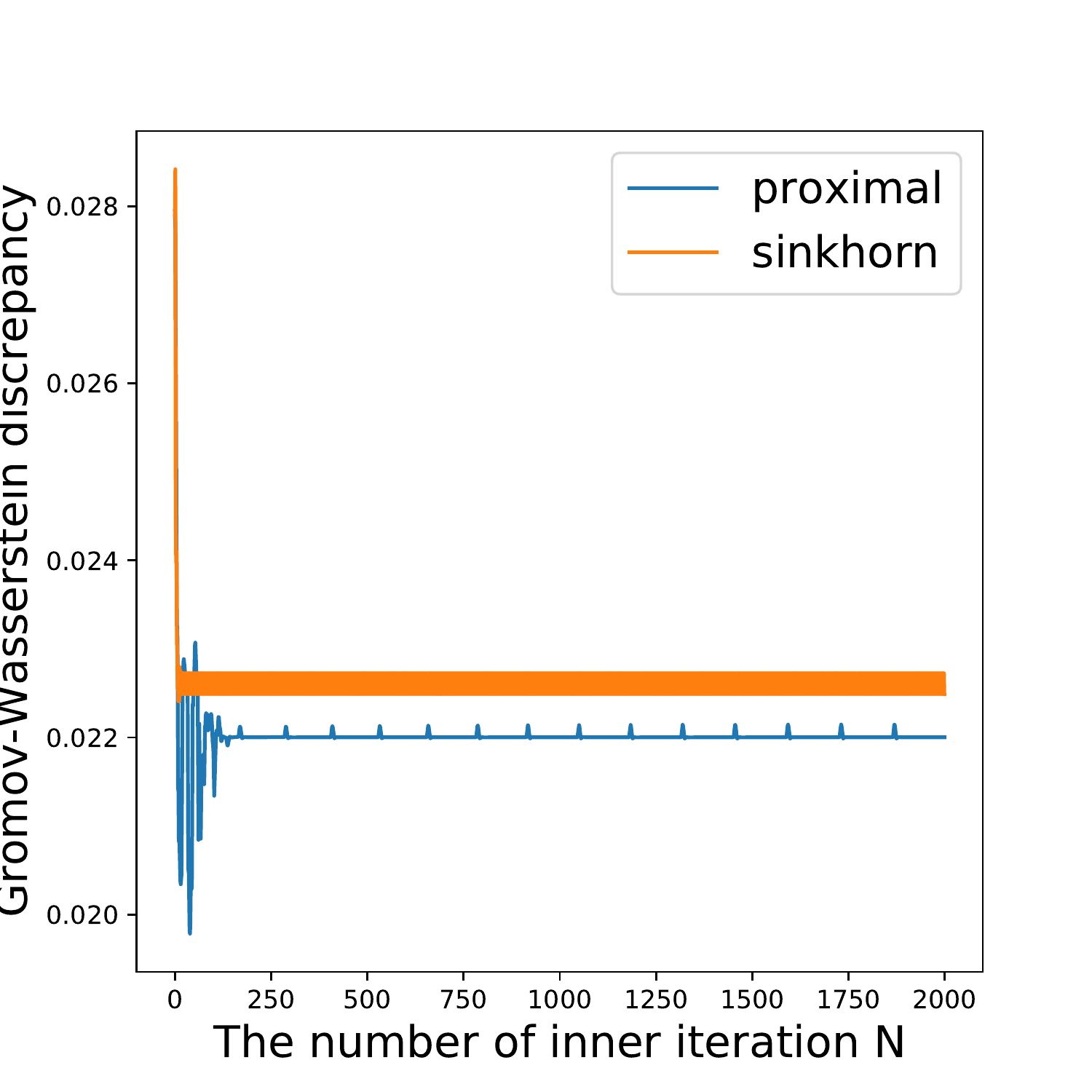}
    }
    \subfigure[$J=100$, $\gamma=1e-2$]{
    \includegraphics[width=3.7cm]{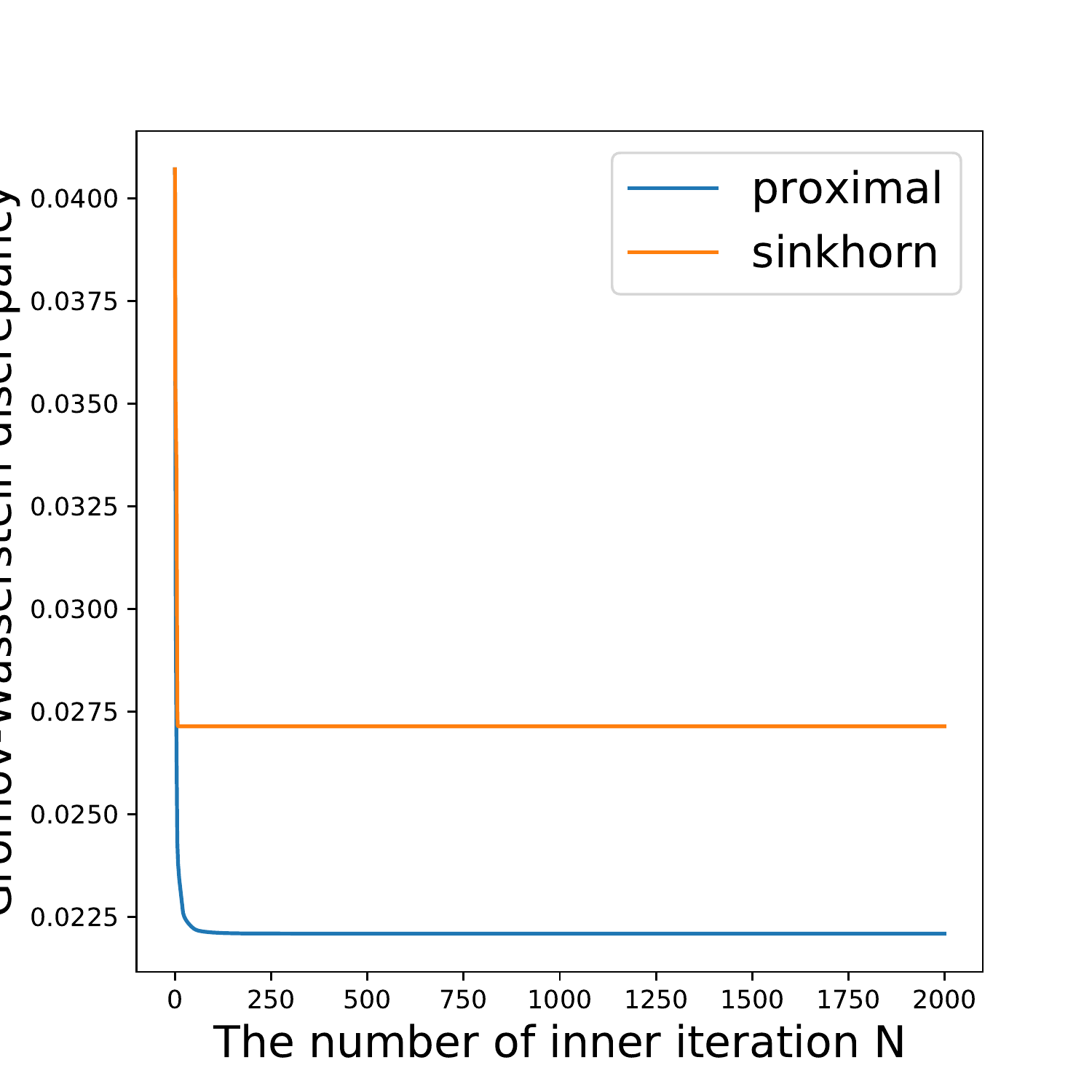}
    }
    \subfigure[$J=100$, $\gamma=1e-1$]{
    \includegraphics[width=3.7cm]{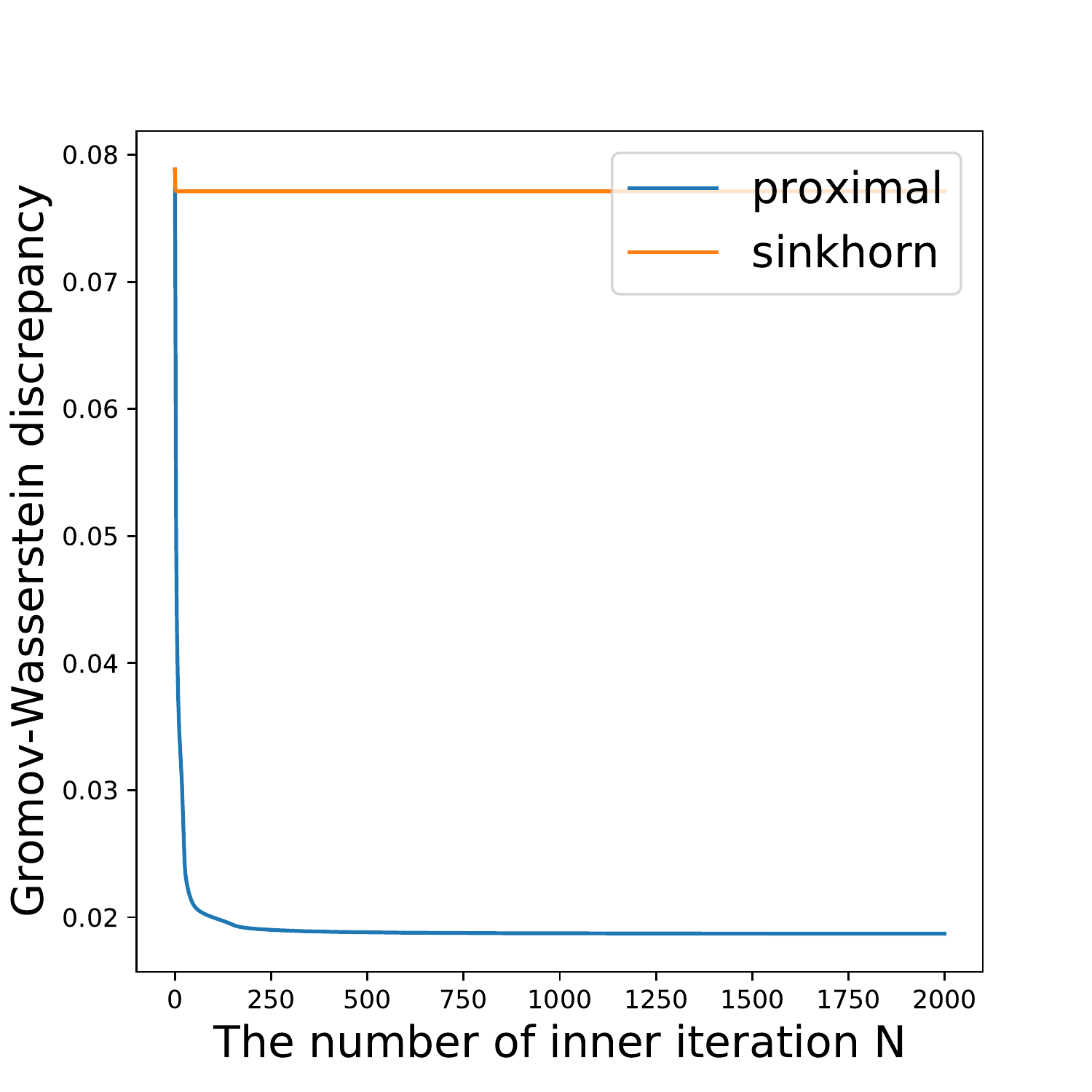}
    }
    \subfigure[$J=100$, $\gamma=1$]{
    \includegraphics[width=3.7cm]{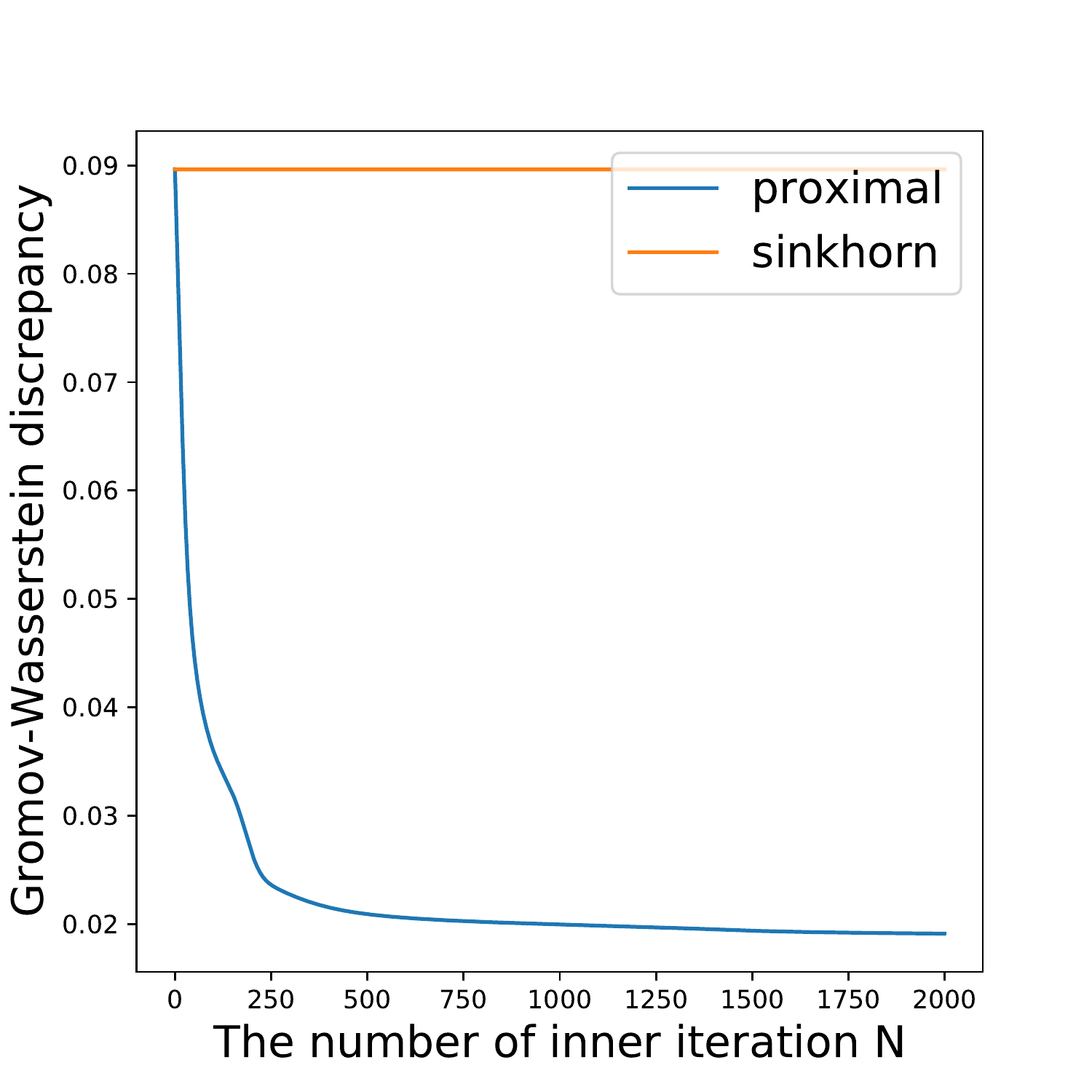}
    }
    \caption{Comparisons for our method (blue curves) and that in~\cite{peyre2016gromov} (orange curves).}
    \label{fig:cmp}
\end{figure*}

\begin{figure*}[t]
    \centering
    \includegraphics[width=1\linewidth]{legend.pdf}
    \subfigure[K-NN: $|\mathcal{V}_s|=50$]{
    \includegraphics[height=3.4cm]{syn_50.pdf}\label{fig:50}
    }
    \subfigure[K-NN: $|\mathcal{V}_s|=100$]{
    \includegraphics[height=3.4cm]{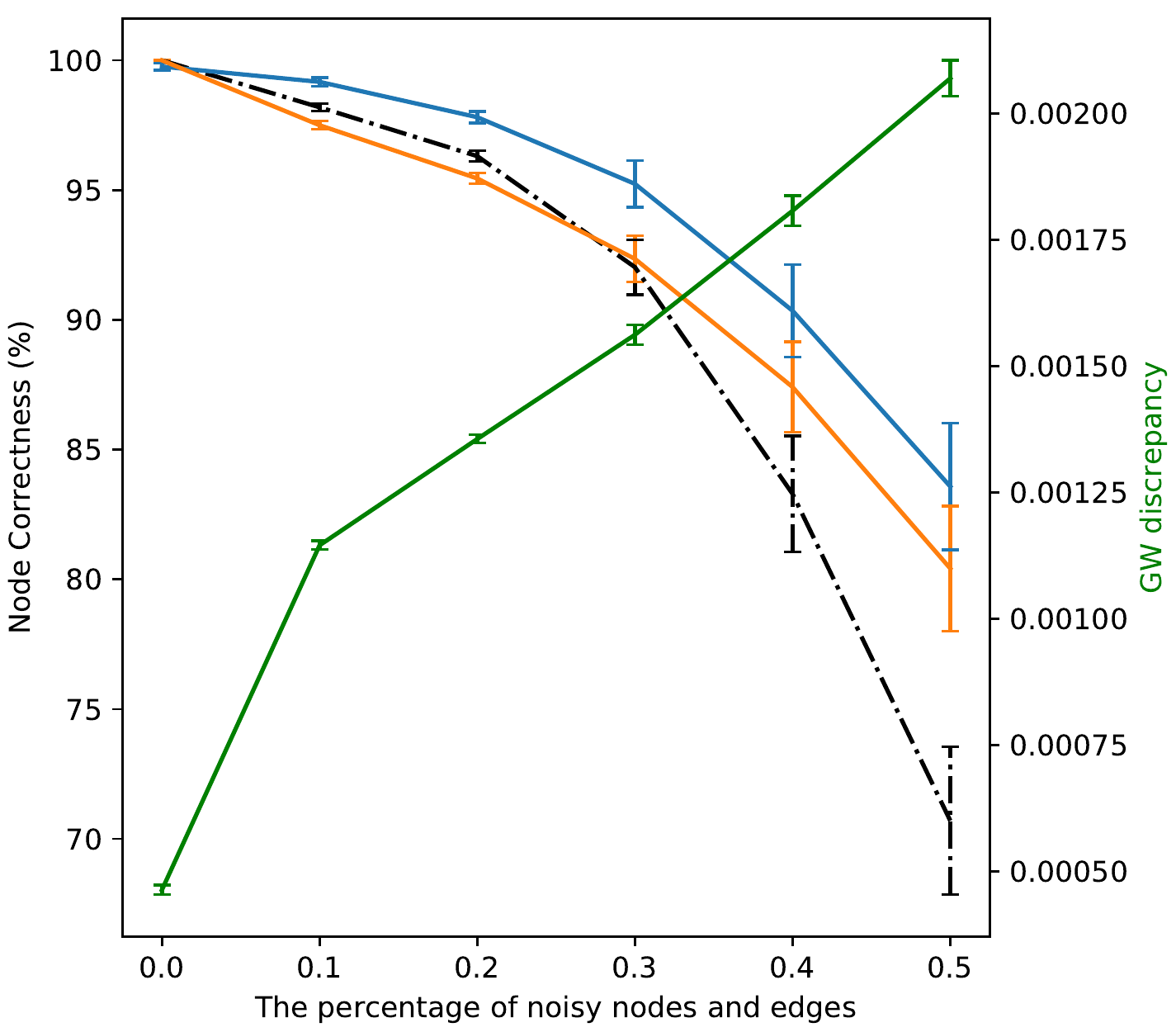}\label{fig:100}
    }
    \subfigure[BA: $|\mathcal{V}_s|=50$]{
    \includegraphics[height=3.4cm]{ba_50.pdf}\label{fig:ba50}
    }
    \subfigure[BA: $|\mathcal{V}_s|=100$]{
    \includegraphics[height=3.4cm]{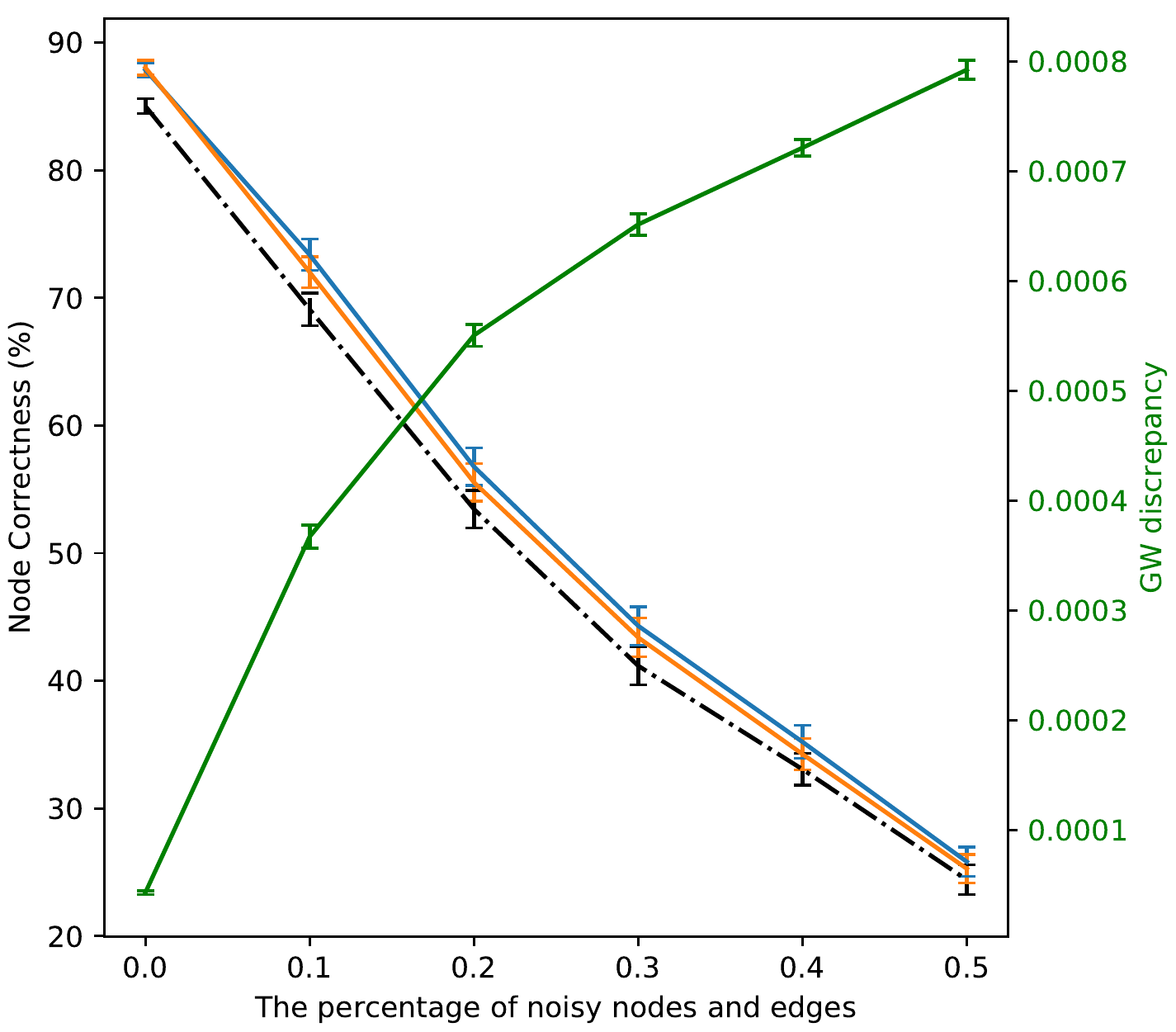}\label{fig:ba100}
    }
    \vspace{-10pt}
    \caption{The performance of our method on synthetic data.}
    \label{fig:syn_full}
\end{figure*}

\subsection{Connections and comparisons with existing method}
Note that when replacing the KL-divergence $\mbox{KL}(\bm{T}\lVert \bm{T}^{(n)})$ in (\ref{eq:ipot}) with an entropy regularizer $H(\bm{T})$, we derive an entropic GW discrepancy, which can also be solved by the Sinkhorn-Knopp algorithm. 
Accordingly, the $\bm{G}$ in Algorithm~\ref{alg1} (line 6) is replaced with $\exp(-\frac{\bm{C}^{(m,n)}}{\gamma})$. 
In such a situation, the proposed algorithm becomes the Sinkhorn projection method in~\cite{peyre2016gromov}.

For both these two methods, the number of Sinkhorn iterations $J$ and the weight of (proximal or entropical) regularizer $\gamma$ are two significant hyperparameters. 
Figure~\ref{fig:cmp} shows the empirical convergence of these two methods using different hyperparameters with respect to the number of inner iterations ($N$ in Algorithm~\ref{alg1}). 
We can find that for the Sinkhorn method it can obtain smaller GW discrepancy than our proximal point method when the weight of regularizer $\gamma$ is very small ($1e-3$) and $J=1$. 
However, in such a situation both of these two methods suffer from a high risk of numerical instability. 
When enlarging $\gamma$, the stability of our method is improved obviously, and it is still able to obtain small GW discrepancy with a good convergence rate. 
The Sinkhorn method, on the contrary, converges slowly when $\gamma>1e-2$. 
In other words, our method is more robust to the change of $\gamma$, and we can choose $\gamma$ in a wide range and achieve a trade-off between convergence and stability easily.
Additionally, although the increase of $J$ helps to improve the stability of our method slightly, $i.e.$, suppressing the numerical fluctuations after the GW discrepancy converges, such an improvement is so obvious as the cost on the computational complexity. 
Therefore, in practice we set $J=1$.

\begin{table*}[t]
\begin{threeparttable}
\caption{Matched disease--procedure pairs based on optimal transport}\label{tab:match}
\tiny{
\centering
\begin{tabular}{
@{\hspace{2pt}}c@{\hspace{2pt}}|
@{\hspace{2pt}}l@{\hspace{2pt}}|
@{\hspace{2pt}}c@{\hspace{2pt}}|
@{\hspace{2pt}}c@{\hspace{2pt}}
}
\hline\hline
$T_{ij}$ &Disease $\leftrightarrow$ Procedure &CR1 &CR2\\
\hline
1.00&d4019: Unspecified essential hypertension $\leftrightarrow$ p9604: Insertion of endotracheal tube 
&
&\\
0.22&d4019: Unspecified essential hypertension $\leftrightarrow$ p966: Enteral infusion of concentrated nutritional substances 
&
&\\
0.17&d4280: Congestive heart failure, unspecified $\leftrightarrow$ p966: Enteral infusion of concentrated nutritional substances 
&\textcolor{red}{\checkmark} 
&\textcolor{red}{\checkmark}\\
0.64&d4280: Congestive heart failure, unspecified $\leftrightarrow$ p9671: Continuous invasive mechanical ventilation for less than 96 consecutive hours 
&\textcolor{red}{\checkmark} 
&\textcolor{red}{\checkmark} \\
0.36&d42731: Atrial fibrillation $\leftrightarrow$ p3961: Extracorporeal circulation auxiliary to open heart surgery 
&\textcolor{blue}{\checkmark} 
&\textcolor{red}{\checkmark} \textcolor{blue}{\checkmark}\\
0.18&d42731: Atrial fibrillation $\leftrightarrow$ p8856: Coronary arteriography using two catheters 
&\textcolor{red}{\checkmark} 
&\textcolor{red}{\checkmark}\textcolor{blue}{\checkmark} \\
0.16&d42731: Atrial fibrillation $\leftrightarrow$ p8872: Diagnostic ultrasound of heart &\textcolor{red}{\checkmark} 
&\textcolor{red}{\checkmark}\\
0.34&d41401: Coronary atherosclerosis of native coronary artery $\leftrightarrow$ p3961: Extracorporeal circulation auxiliary to open heart surgery 
&\textcolor{red}{\checkmark} 
&\textcolor{red}{\checkmark}\\
0.29&d41401: Coronary atherosclerosis of native coronary artery $\leftrightarrow$ p8856: Coronary arteriography using two catheters 
&\textcolor{red}{\checkmark} 
&\textcolor{red}{\checkmark}\\
0.42&d5849: Acute kidney failure, unspecified $\leftrightarrow$ p9672: Continuous invasive mechanical ventilation for 96 consecutive hours or more 
&
&\textcolor{blue}{\checkmark}\\
0.44&d25000: Diabetes mellitus without mention of complication, type II or unspecified type, not stated as uncontrolled $\leftrightarrow$ p3615: Single internal mammary-coronary artery bypass 
&\textcolor{red}{\checkmark}\tnote{a}
&\\
0.45&d2724 Other and unspecified hyperlipidemia $\leftrightarrow$ p8853 Angiocardiography of left heart structures
&\textcolor{red}{\checkmark} 
&\\
0.20&d51881: Acute respiratory failure $\leftrightarrow$ p3893: Venous catheterization, not elsewhere classified 
&\textcolor{red}{\checkmark} 
&\textcolor{red}{\checkmark}\\
0.22&d51881: Acute respiratory failure $\leftrightarrow$ p9904: Transfusion of packed cells 
&\textcolor{blue}{\checkmark}
&\textcolor{red}{\checkmark}\textcolor{blue}{\checkmark}\\
0.29&d5990: Urinary tract infection, site not specified $\leftrightarrow$ p3893: Venous catheterization, not elsewhere classified 
&\textcolor{red}{\checkmark} 
&\textcolor{red}{\checkmark}\\
0.22&d53081: Esophageal reflux $\leftrightarrow$ p9390: Non-invasive mechanical ventilation 
&
&\textcolor{blue}{\checkmark}\\
0.23&d2720: Pure hypercholesterolemia $\leftrightarrow$ p3891: Arterial catheterization &&\textcolor{red}{\checkmark} \\
0.48&dV053: Need for prophylactic vaccination and inoculation against viral hepatitis $\leftrightarrow$ p9955: Prophylactic administration of vaccine against other diseases 
&\textcolor{red}{\checkmark} 
&\textcolor{red}{\checkmark}\\
0.53&dV290: Observation for suspected infectious condition $\leftrightarrow$ p9955: Prophylactic administration of vaccine against other diseases 
&\textcolor{red}{\checkmark} 
&\\
0.30&d2859: Anemia, unspecified $\leftrightarrow$ p9915: Parenteral infusion of concentrated nutritional substances 
&\textcolor{red}{\checkmark}\tnote{b}
&\\
0.25&d2449: Unspecified acquired hypothyroidism $\leftrightarrow$ p9915: Parenteral infusion of concentrated nutritional substances 
&
&\\
0.24&d486: Pneumonia, organism unspecified $\leftrightarrow$ p9671: Continuous invasive mechanical ventilation for less than 96 consecutive hours 
&\textcolor{red}{\checkmark}\textcolor{blue}{\checkmark} 
&\textcolor{red}{\checkmark}\textcolor{blue}{\checkmark}\\
0.18&d2851: Acute posthemorrhagic anemia $\leftrightarrow$ p9904: Transfusion of packed cells 
&\textcolor{red}{\checkmark} 
&\textcolor{red}{\checkmark}\\
0.18&d2762: Acidosis $\leftrightarrow$ p966: Enteral infusion of concentrated nutritional substances 
&
&\textcolor{red}{\checkmark}\textcolor{blue}{\checkmark} \\
0.28&d496: Chronic airway obstruction, not elsewhere classified $\leftrightarrow$ p3722: Left heart cardiac catheterization 
&
&\\
0.16&d99592: Severe sepsis $\leftrightarrow$ p3893: Venous catheterization, not elsewhere classified 
&\textcolor{red}{\checkmark} 
&\textcolor{red}{\checkmark}\\
0.26&d0389: Unspecified septicemia $\leftrightarrow$ p966: Enteral infusion of concentrated nutritional substances 
&\textcolor{blue}{\checkmark} 
&\\
0.26&d5070: Pneumonitis due to inhalation of food or vomitus $\leftrightarrow$ p3893: Venous catheterization, not elsewhere classified 
&\textcolor{red}{\checkmark} 
&\textcolor{red}{\checkmark}\\
0.33&dV3000: Single liveborn, born in hospital, delivered without mention of cesarean section $\leftrightarrow$ p331: Incision of lung 
&
&\\
0.17&d5859: Chronic kidney disease, unspecified $\leftrightarrow$ p9904: Transfusion of packed cells &\textcolor{red}{\checkmark} &\textcolor{red}{\checkmark}\\
0.13&d311: Depressive disorder - not elsewhere classified $\leftrightarrow$ p4513: Other endoscopy of small intestine
&
&\\
0.14&d40390: Hypertensive chronic kidney disease $\leftrightarrow$ p3324: Closed biopsy of bronchus
&
&\\
0.11&d3051: Tobacco use disorder $\leftrightarrow$ p5491: Percutaneous abdominal drainage
&
&\\
0.16&d412: Old myocardial infarction $\leftrightarrow$ p8853: Angiocardiography of left heart structures &\textcolor{red}{\checkmark} &\textcolor{red}{\checkmark}\\
0.18&d2875: Thrombocytopenia, unspecified $\leftrightarrow$ p3893: Venous catheterization, not elsewhere classified 
&\textcolor{red}{\checkmark} 
&\textcolor{red}{\checkmark}\\
0.10&dV4581: Aortocoronary bypass status $\leftrightarrow$ p3891: Arterial catheterization 
&
&\textcolor{red}{\checkmark} \\
0.25&d41071: Subendocardial infarction, initial episode of care $\leftrightarrow$ p3723: Combined right and left heart cardiac catheterization 
&\textcolor{red}{\checkmark}\textcolor{blue}{\checkmark} 
&\textcolor{red}{\checkmark}\textcolor{blue}{\checkmark} \\
0.09&d2761: Hyposmolality and/or hyponatremia $\leftrightarrow$ p966: Enteral infusion of concentrated nutritional substances
&
&\textcolor{red}{\checkmark}\textcolor{blue}{\checkmark} \\
0.21&d4240: Mitral valve disorders $\leftrightarrow$ p9904: Transfusion of packed cells &&\textcolor{red}{\checkmark} \\
0.31&dV3001: Single liveborn, born in hospital, delivered by cesarean section $\rightarrow$ p640: Circumcision 
&\textcolor{red}{\checkmark} 
&\textcolor{red}{\checkmark}\\
0.08&d5119: Unspecified pleural effusion $\leftrightarrow$ p9604: Insertion of endotracheal tube
&
&\textcolor{red}{\checkmark} \\
0.07&dV4582: Percutaneous transluminal coronary angioplasty status $\leftrightarrow$ p9907: Transfusion of other serum
&\textcolor{red}{\checkmark} 
&\\
0.23&d40391: Hypertensive chronic kidney disease, unspecified, with chronic kidney disease stage V or end stage renal disease $\leftrightarrow$ p3893: Venous catheterization, not elsewhere classified 
&\textcolor{red}{\checkmark} 
&\textcolor{red}{\checkmark}\\
0.17&d78552: Septic shock $\leftrightarrow$ p9904: Transfusion of packed cells 
&
&\textcolor{red}{\checkmark}\\
0.05&d4241: Aortic valve disorders $\leftrightarrow$ p8872: Diagnostic ultrasound of heart
&\textcolor{red}{\checkmark} 
&\textcolor{red}{\checkmark} \\
0.06&dV5867: Long-term (current) use of insulin $\leftrightarrow$ p3995: Hemodialysis
&
&\\
0.07&d42789: Other specified cardiac dysrhythmias $\leftrightarrow$ p9604: Insertion of endotracheal tube
&
&\\
0.05&d32723: Obstructive sleep apnea (adult)(pediatric) $\leftrightarrow$ p9390: Non-invasive mechanical ventilation
&\textcolor{blue}{\checkmark} 
&\textcolor{red}{\checkmark}\\
0.17&d9971: Cardiac complications, not elsewhere classified $\leftrightarrow$ p8856: Coronary arteriography using two catheters 
&\textcolor{red}{\checkmark}\textcolor{blue}{\checkmark}  
&\textcolor{red}{\checkmark}\textcolor{blue}{\checkmark} \\
0.07&d5845: Acute kidney failure with lesion of tubular necrosis $\leftrightarrow$ p9904: Transfusion of packed cells
&\textcolor{blue}{\checkmark} 
&\textcolor{red}{\checkmark}\textcolor{blue}{\checkmark} \\
0.05&d2760: Hyperosmolality and/or hypernatremia $\leftrightarrow$ p966: Enteral infusion of concentrated nutritional substances
&
&\textcolor{red}{\checkmark}\textcolor{blue}{\checkmark} \\
0.27&d7742: Neonatal jaundice associated with preterm delivery $\leftrightarrow$ p9983: Other phototherapy 
&\textcolor{red}{\checkmark}
&\textcolor{red}{\checkmark}\\
0.12&d49390: Asthma - unspecified type - unspecified $\leftrightarrow$ p5491: Percutaneous abdominal drainage
&
&\\
0.10&d2767: Hyperpotassemia $\leftrightarrow$ p3893: Venous catheterization - not elsewhere classified
&\textcolor{red}{\checkmark} 
&\textcolor{red}{\checkmark} \\
0.09&d5180: Pulmonary collapse $\leftrightarrow$ p3893: Venous catheterization - not elsewhere classified
&\textcolor{red}{\checkmark} 
&\textcolor{red}{\checkmark} \\
0.08&d4168: Other chronic pulmonary heart diseases $\leftrightarrow$ p9907: Transfusion of other serum
&\textcolor{red}{\checkmark} 
&\\
0.13&d45829: Other iatrogenic hypotension $\leftrightarrow$ p9907: Transfusion of other serum
&\textcolor{red}{\checkmark} 
&\\
0.14&d2749: Gout - unspecified $\leftrightarrow$ p3995: Hemodialysis
&\textcolor{red}{\checkmark} 
&\\
0.10&d4589: Hypotension - unspecified $\leftrightarrow$ p966: Enteral infusion of concentrated nutritional substances
&
&\\
0.25&dV502: Routine or ritual circumcision $\leftrightarrow$ p9955: Prophylactic administration of vaccine against other diseases 
&\textcolor{red}{\checkmark}\tnote{c}
&\\
\hline\hline
\end{tabular}
\begin{tablenotes}
 \item[a] The relationship here is that usually people with diabetes also have heart disease, and heart disease can require a coronary artery bypass
 \item[b] The relationship here is that if someone has a chronic disease  they can develop anemia of chronic disease and they may also be requiring parenteral nutrition for some specific condition
 \item[c] This procedure is not inherently related to the disease, but they do appear together frequently in the same medical record because they both happen to newborn babies.
 \item[\textcolor{red}{\checkmark}] The procedure is related to the treatment of the disease.
 \item[\textcolor{blue}{\checkmark}] The procedure can lead to the disease as side effect or complication.
\end{tablenotes}
}
\end{threeparttable}
\end{table*}

\subsection{Runtime}
We implement our method based on PyTorch. 
For the synthetic graphs with 100 nodes, it takes about 15 seconds on a CPU. 
For the MC3 graphs (622 nodes), it takes about 8 minutes on the CPU. 
It is even faster than some baselines based on C++~\cite{vijayan2015magna++,kuchaiev2011integrative}.

\subsection{More details of experiments}
For synthetic data, Figure~\ref{fig:syn_full} shows all experimental results on node correctness. 
We can find that the proposed method consistently better than the baseline method.

\begin{figure*}[t]
    \centering
    \includegraphics[width=10cm]{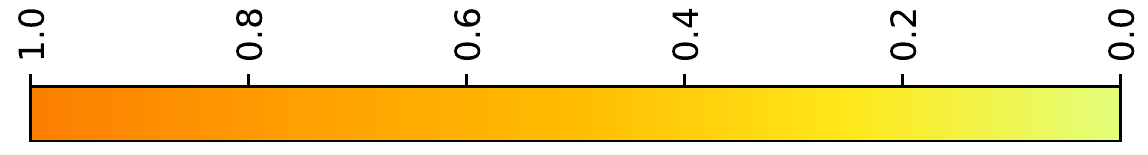}
    \includegraphics[width=10cm]{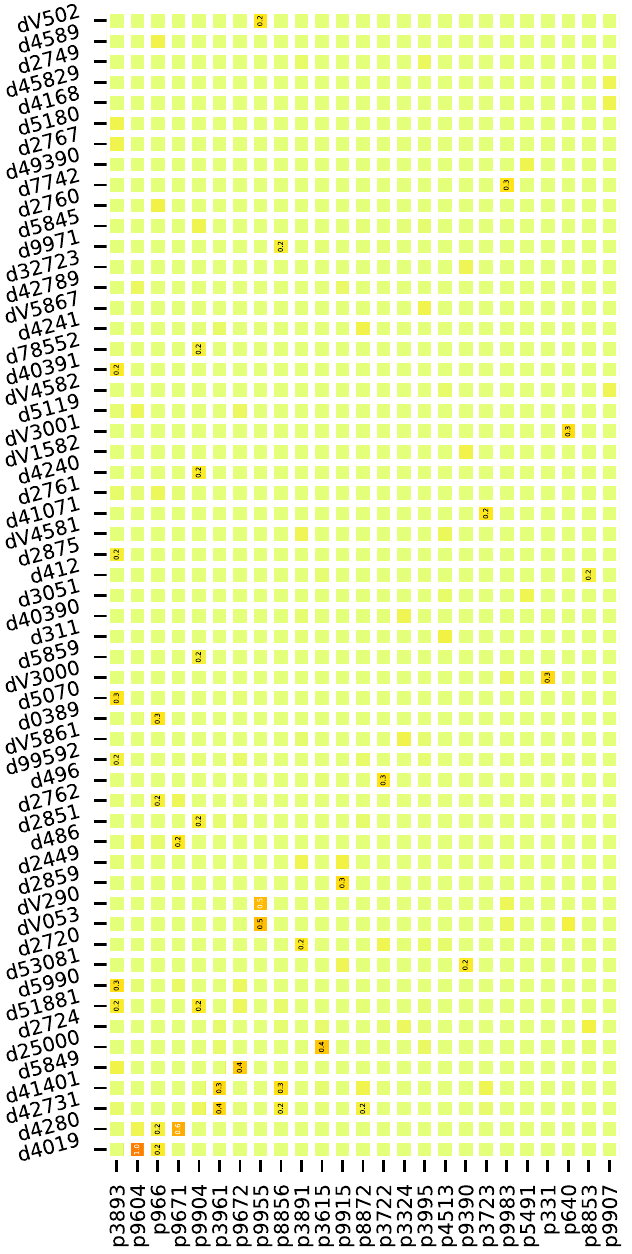}
    \caption{The optimal transport from diseases to procedures (enlarged version).}
    \label{fig:ot}
\end{figure*}

For the MIMIC-III data, the enlarged optimal transport between diseases and procedures learned by our method is shown in Figure~\ref{fig:ot}. 
The transport matrix is normalized, whose maximum value is 1.
For each disease, the procedures with the maximum transport value or the corresponding $\widehat{T}_{ij}\geq 0.15$ are listed in Table~\ref{tab:match}.
Additionally, we ask two clinical researchers to evaluate these pairs --- for each pair, each research independently checks whether the procedure is potentially related to the disease. 
The columns of ``CR1'' and ``CR2'' in Table~\ref{tab:match} give the evaluation results. 
For each pair, the ``\textcolor{red}{\checkmark}'' means that the procedure is potentially related to the treatments of the disease, while the ``\textcolor{blue}{\checkmark}'' means that the procedure can lead to the disease as side effect or complication.
We can find that 1) the evaluation results from different clinical researchers are with high consistency; 2) over $73.6\%$ of the pairs are reasonable: they correspond to either ``diseases and their treatments'' or ``procedures and their complications''. 
These phenomena demonstrate that the learned optimal transport is clinically-meaningful to some extent, which reflects some relationships between diseases and procedures.
Table~\ref{tab:icd_map} lists the ICD codes of diseases and procedures and their detailed descriptions.

\begin{table*}[!t]
\caption{The map between ICD codes and diseases/procedures}
\centering
\tiny{
\setlength{\tabcolsep}{5pt}
\begin{tabular}{@{\hspace{2pt}}l@{\hspace{2pt}}
@{\hspace{2pt}}l@{\hspace{2pt}}
} 
\hline\hline
ICD code & Disease/Procedure\\
\hline
d4019&Unspecified essential hypertension\\
d41401&Coronary atherosclerosis of native coronary artery\\
d4241&Aortic valve disorders\\
dV4582&Percutaneous transluminal coronary angioplasty status\\
d2724&Other and unspecified hyperlipidemia\\
d486&Pneumonia, organism unspecified\\
d99592&Severe sepsis\\
d51881&Acute respiratory failure\\
d5990&Urinary tract infection, site not specified\\
d5849&Acute kidney failure, unspecified\\
d78552&Septic shock\\
d25000&Diabetes mellitus without mention of complication, type II or unspecified type\\
d2449&Unspecified acquired hypothyroidism\\
d41071&Subendocardial infarction, initial episode of care\\
d4280&Congestive heart failure, unspecified\\
d4168&Other chronic pulmonary heart diseases\\
d412&Pneumococcus infection in conditions classified elsewhere and of unspecified site\\
d2761&Hyposmolality and/or hyponatremia\\
d2720&Pure hypercholesterolemia\\
d2762&Acidosis\\
d389&Unspecified septicemia\\
d4589&Hypotension, unspecified\\
d42731&Atrial fibrillation\\
d2859&Anemia, unspecified\\
d311&Cutaneous diseases due to other mycobacteria\\
dV3001&Single liveborn, born in hospital, delivered by cesarean section\\
dV053&Need for prophylactic vaccination and inoculation against viral hepatitis\\
d4240&Mitral valve disorders\\
dV3000&Single liveborn, born in hospital, delivered without mention of cesarean section\\
d7742&Neonatal jaundice associated with preterm delivery\\
d42789&Other specified cardiac dysrhythmias\\
d5070&Pneumonitis due to inhalation of food or vomitus\\
dV502&Routine or ritual circumcision\\
d2760&Hyperosmolality and/or hypernatremia\\
dV1582&Personal history of tobacco use\\
d40390&Hypertensive chronic kidney disease, unspecified, with chronic kidney disease stage I through stage IV, or unspecified\\
dV4581&Aortocoronary bypass status\\
dV290&Observation for suspected infectious condition\\
d5845&Acute kidney failure with lesion of tubular necrosis\\
d2875&Thrombocytopenia, unspecified\\
d2767&Hyperpotassemia\\
d32723&Obstructive sleep apnea (adult)(pediatric)\\
dV5861&Long-term (current) use of anticoagulants\\
d2851&Acute posthemorrhagic anemia\\
d53081&Esophageal reflux\\
d496&Chronic airway obstruction, not elsewhere classified\\
d40391&Hypertensive chronic kidney disease, unspecified, with chronic kidney disease stage V or end stage renal disease\\
d9971&Gross hematuria\\
d5119&Unspecified pleural effusion\\
d2749&Gout, unspecified\\
d5859&Chronic kidney disease, unspecified\\
d49390&Asthma, unspecified type, unspecified\\
d45829&Other iatrogenic hypotension\\
d3051&Tobacco use disorder\\
dV5867&Long-term (current) use of insulin\\
d5180&Pulmonary collapse\\ \hline
p9604&Insertion of endotracheal tube\\
p9671&Continuous invasive mechanical ventilation for less than 96 consecutive hours\\
p3615&Single internal mammary-coronary artery bypass\\
p3961&Extracorporeal circulation auxiliary to open heart surgery\\
p8872&Diagnostic ultrasound of heart\\
p9904&Transfusion of packed cells\\
p9907&Transfusion of other serum\\
p9672&Continuous invasive mechanical ventilation for 96 consecutive hours or more\\
p331&Spinal tap\\
p3893&Venous catheterization, not elsewhere classified\\
p966&Enteral infusion of concentrated nutritional substances\\
p3995&Hemodialysis\\
p9915&Parenteral infusion of concentrated nutritional substances\\
p8856&Coronary arteriography using two catheters\\
p9955&Prophylactic administration of vaccine against other diseases\\
p3891&Arterial catheterization\\
p9390&Non-invasive mechanical ventilation\\
p9983&Other phototherapy\\
p640&Circumcision\\
p3722&Left heart cardiac catheterization\\
p8853&Angiocardiography of left heart structures\\
p3723&Combined right and left heart cardiac catheterization\\
p5491&Percutaneous abdominal drainage\\
p3324&Closed (endoscopic) biopsy of bronchus\\
p4513&Other endoscopy of small intestine\\
\hline\hline
\end{tabular}\label{tab:icd_map}
}
\end{table*}

\end{document}